\documentclass[journal]{IEEEtran}
\ifCLASSINFOpdf
\else
\fi
\usepackage{amsmath,amssymb,amsfonts}
\usepackage{caption}
\usepackage{newtxmath}
\usepackage{algorithmic}
\usepackage{graphicx}
\usepackage{textcomp}
\usepackage{xcolor}
\usepackage{multirow}
\usepackage{subfigure}
\usepackage{balance}
\usepackage{booktabs}
\usepackage{threeparttable}
\usepackage{url} 
\usepackage[ruled,linesnumbered]{algorithm2e}
\newcommand{\nosection}[1]{\vspace{2pt}\noindent\textbf{#1.}}

\newcommand{\M}{WassFFed}
\newcommand{\MLocal}{WassFFed-local}

\newtheorem{definition}{Definition}
\begin{document}
%
\title{WassFFed: Wasserstein Fair Federated Learning}
%
%
%

\author{Zhongxuan Han, Li Zhang, Chaochao Chen, \IEEEmembership{Senior Member,~IEEE}, Xiaolin Zheng, \IEEEmembership{Senior Member,~IEEE}, Fei Zheng, Yuyuan Li, Jianwei Yin, \IEEEmembership{Senior Member,~IEEE}
\thanks{Zhongxuan Han, Chaochao Chen, Xiaolin Zheng, Fei Zheng, and Jianwei Yin was with the College of Computer Science, Zhejiang University, Hangzhou,
Zhejiang 310027, China. E-mail: \{zxhan, zjuccc, xlzheng, zfscgy2, zjuyjw\}@cs.zju.edu.cn}
\thanks{Li Zhang was with the Polytechnic Institute, Zhejiang University, Hangzhou, Zhejiang 310027, China. E-mail: zhanglizl80@gmail.com.}
\thanks{Yuyuan Li was with the School of Communication Engineering, Hangzhou Dianzi University, Hangzhou, Zhejiang 310027, China. E-mail: y2li@hdu.edu.cn.}
\thanks{Chaochao Chen is the corresponding author.}
}

%
%

\markboth{Journal of \LaTeX\ Class Files,~Vol.~14, No.~8, August~2015}%
{Shell \MakeLowercase{\textit{et al.}}: Bare Demo of IEEEtran.cls for IEEE Journals}
%


\maketitle




%
\IEEEpeerreviewmaketitle
\begin{abstract}
Federated Learning (FL) employs a training approach to address scenarios where users' data cannot be shared across clients.
Achieving fairness in FL is imperative since training data in FL is inherently geographically distributed among diverse user groups.
Existing research on fairness predominantly assumes
access to the entire training data, making direct transfer to FL challenging.
However, the limited existing research on fairness in FL does not effectively address two key challenges, i.e., \textbf{(CH1)} Current methods fail to deal with the inconsistency between fair optimization results obtained with surrogate functions and fair classification results.
\textbf{(CH2)} Directly aggregating local fair models does not always yield a globally fair model due to non-Identical and Independent data Distributions (non-IID) among clients.
To address these challenges, we propose a \textbf{Wass}erstein \textbf{F}air \textbf{Fed}erated Learning framework, namely WassFFed.
\textit{To tackle CH1}, we ensure that the outputs of local models, rather than the loss calculated with surrogate functions or classification results with a threshold, remain independent of various user groups.
\textit{To resolve CH2}, we employ a Wasserstein barycenter calculation of all local models' outputs for each user group, bringing local model outputs closer to the global output distribution to ensure consistency between the global model and local models.
We conduct extensive experiments on three real-world datasets, demonstrating that \M~outperforms existing approaches in striking a balance between accuracy and fairness.
%
\end{abstract}


\section{introduction \label{sec:introduction}}
Fairness has recently become an essential part of the Machine Learning (ML) community~\cite{binns2018fairness,han2023processing,hutchinson201950,mehrabi2021survey}.
Various fairness notions have been proposed in the past few years~\cite{dwork2012fairness,hardt2016equality,zafar2017parity,binns2018fairness}.
Among them, \textit{Group Fairness}~\cite{hardt2016equality,zafar2017parity} stands as one of the most extensively explored notions, emphasizing equal treatment of distinct user groups by ML models.
Most research on fairness in ML assumes access to the entire training data, i.e., centralized learning.
However, in many practical applications, users' data is distributed across different platforms or clients and cannot be shared due to privacy concerns.
This constraint significantly constrains the effectiveness of traditional centralized fair ML models.

To mitigate the necessity of sharing users' data across clients, Federated Learning (FL)~\cite{mcmahan2017communication,smith2017federated,yang2019federated} has emerged as a promising solution.
FL employs a training approach, wherein local models are trained on localized data samples, and their parameters are aggregated to construct a global model.
It is essential to achieve fairness in FL since the training data in FL is always geo-distributed~\cite{du2021fairness} among various groups.
This paper concentrates on the attainment of group fairness within the context of FL.

\begin{figure}[t]
    \centering
    \subfigure[A fair classification model]{
		\label{introduction-centralized-fair}
		\includegraphics[width=0.23\linewidth]{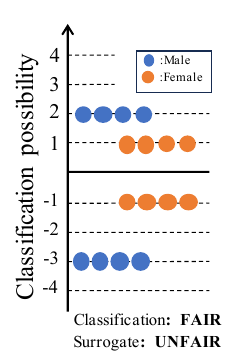}}
    \subfigure[An unfair classification model]{
            \label{introduction-centralized-unfair}
		\includegraphics[width=0.22\linewidth]{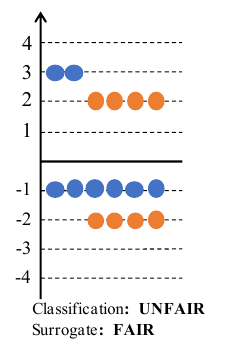}
		}
     \subfigure[Fair local classification models]{
            \label{introduction-local-fair}
		\includegraphics[width=0.2\linewidth]{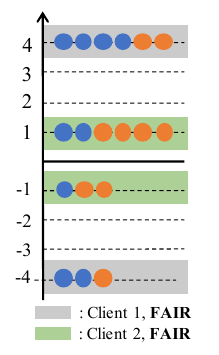}
		} 
    \subfigure[An unfair global classification model]{
            \label{introduction-global-unfair}
		\includegraphics[width=0.2\linewidth]{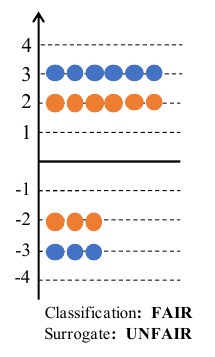}
		} 
    \caption{The samples above the horizontal solid line are predicted to be positive samples and vice versa. (a) visualizes a fair classification but is considered unfair according to the surrogate function ($male = -0.5$, $female = 0$). (b) visualizes an unfair classification model but is considered fair according to the surrogate function ($male = 0$, $female = 0$). (c) visualizes two local fair classification models that are also considered fair with the surrogate function. (d) depicts the global model derived from aggregating the two models in (c). Upon aggregation, this global model yields fair classification results; however, when evaluated with the surrogate function, it is regarded as unfair ($male = 1$, $female = 0.67$).
    }
    \label{introduction-figure}
\end{figure}

Existing research on fairness in FL is limited and fails to tackle two key challenges.
\textbf{CH1:} \textit{Current methods fail to deal with the inconsistency between fair optimization results obtained with surrogate functions and fair classification results.}
Many researchers treat the training of fair classification models as a constraint optimization problem~\cite{han2023processing,goh2016satisfying,menon2018cost} by minimizing a loss function subject to certain fairness constraints, e.g., demographic parity~\cite{dwork2012fairness} and equal opportunity~\cite{hardt2016equality}.
However, most quantitative fairness metrics are non-convex due to the use of indicator functions (i.e., $\vmathbb{1}(x) \in \{0, 1\}$), rendering the optimization problem intractable.
A widely adopted strategy is to employ surrogate functions that resemble indicator functions while maintaining continuity and convexity to address fair optimization challenges~\cite{goh2016satisfying,zafar2017fairness}.
Nevertheless, given the inherent differences between surrogate functions and the original non-convex indicator function, estimation errors are inevitable~\cite{wu2019convexity,lohaus2020too}.
Figure~\ref{introduction-centralized-fair} and~\ref{introduction-centralized-unfair} provide two examples that illustrate the inconsistency between the estimation of surrogate functions and classification results, \textit{where the surrogate function evaluates the average classification possibility for each group.}
Such inconsistencies can lead the optimization process astray, resulting in unfair classification outcomes.
\textbf{FL aggregates local models to construct a global model, introducing unique estimation errors.}
As illustrated in Figure~\ref{introduction-local-fair}, two clients both build fair classification models.
However, the aggregated global model, as shown in Figure~\ref{introduction-global-unfair}, continues to yield fair classification results, yet it is considered unfair according to the surrogate function.
The phenomenon proves that FL may result in estimation inconsistencies between surrogate functions and real classification results, even though this issue is not universally present across all clients.
Existing fair FL methods ignore this problem, thereby limiting the effectiveness of fair optimization results and introducing instability in the training process.

\textbf{CH2:} \textit{Existing research typically involves directly aggregating local fair models. However, this approach does not consistently yield a globally fair model due to non-Identical and Independent data Distributions (non-IID) among clients.}
Most of the currently proposed methods aim to enhance the fairness of global models by mitigating bias in local models~\cite{abay2020mitigating,ezzeldin2023fairfed}, i.e., Locally Fair Training (LFT).
However, the data distribution among clients consistently exhibits non-IID characteristics, particularly when users possess different sensitive attributes.
Aggregating local fair models doesn't always guarantee a fair global model~\cite{wang2023mitigating}.
For example, consider a multinational bank seeking to create a global model by aggregating local models trained in different countries.
Given that users in different countries often belong to diverse racial backgrounds, the aggregated global model may face severe unfairness in race, even when all local models individually achieve fairness.
Recently, FEDFB~\cite{zeng2021improving} attempts to address this challenge by calculating coefficients for local fairness constraints at the server level.
However, due to the intricate data distributions across all clients, solely modifying these coefficients fails to yield substantial improvements.

In this paper, we propose a novel \textbf{Wass}erstein \textbf{F}air \textbf{Fed}erated Learning framework, namely~\M, to tackle the aforementioned challenges.
Generally, \M~calculates the Wasserstein barycenter\cite{vallender1974calculation,jiang2020wasserstein} among the distributions of model outputs corresponding to groups of users with different sensitive attributes.
Subsequently, \M~imposes a small Wasserstein distance between these distributions and the computed barycenter.
This process encourages similarity among model outputs for users with diverse sensitive attributes, ultimately promoting fairness in model predictions.
In detail, to tackle \textbf{CH1}, we directly concentrate on the outputs of the classification models, instead of calculating a fairness loss based on surrogate functions.
Therefore, we can avoid the estimation error caused by the user of surrogate functions.
Since classification models invariably produce continuous outputs, with classification results determined by thresholds,
ensuring that model outputs remain independent of sensitive attributes guarantees fairness in classification outcomes, regardless of threshold variations.
To tackle \textbf{CH2}, \M~computes the Wasserstein barycenter on the server, drawing from the distributions of outputs from all local models.
To achieve this, all clients share their model output distributions with the server.
Then, the server aggregates the received distributions to construct global model output distributions, each corresponding to users with distinct sensitive attributes.
Following this aggregation, the server proceeds to calculate a global Wasserstein barycenter grounded in these distributions.
Finally, \M~enforces distributions of all clients' outputs corresponding to users with different sensitive attributes to be closer to the global barycenter.
By doing so, we avoid the essential fairness inconsistency between the global model and local models caused by non-IID data distributions.
Since all local models will share the same output distribution, similar to the global Wasserstein barycenter.

We conduct extensive experiments on three publicly available real-world datasets, compared with State-Of-The-Art (SOTA) methods.
The experimental results conclusively demonstrate that our proposed \M~outperforms existing methods, achieving a superior balance between accuracy and fairness.
Notably, \M~consistently excels in more complex classification tasks, showcasing its remarkable generalizability.

We summarize our main contributions as follows:
\begin{enumerate}
    \item We introduce a novel framework named \M~to effectively address fairness issues in FL.
    \item Our approach avoids the inherent estimation errors associated with training a fair model using surrogate models and attains consistent fairness results across both global and local models.
    \item We conduct extensive experiments on three publicly available real-world datasets to demonstrate the efficiency of the proposed \M~framework.
\end{enumerate}

\section{Related Work}
This paper focuses on tackling the fairness issue in federated learning, therefore, we introduce the related work in two parts, fairness in machine learning and fair federated learning.

\subsection{Fairness in Machine Learning}
Fairness has garnered significant attention due to the growing deployment of ML systems in real-world scenarios.
As presented in \cite{mehrabi2021survey}, fairness, in the context of decision-making processes, is broadly defined as the absence of any prejudice or favoritism toward an individual or a group based on their inherent or acquired characteristics. 
From various perspectives, research in ML fairness can be categorized into different domains.

\textit{Concerning the groups affected by fairness issues}, research on fairness can be categorized into two primary aspects: group fairness and individual fairness \cite{mehrabi2021survey, dai2022comprehensive}.
Group fairness seeks to ensure equal treatment for users from different groups. 
Notable approaches in this domain include Equalized Odds \cite{hardt2016equality}, Equal Opportunity \cite{hardt2016equality}, Conditional Statistical Parity \cite{corbett2017algorithmic}, Demographic Parity \cite{kusner2017counterfactual}, and Treatment Equality \cite{berk2021fairness}.
Individual fairness aims to provide similar individuals with similar recommendation results. 
Relevant research in this area encompasses Fairness Through Unawareness \cite{grgic2016case}, Fairness Through Awareness \cite{dwork2012fairness}, and Counterfactual Fairness~\cite{kusner2017counterfactual}.

\textit{Concerning different stages of the ML process that the fairness algorithms are applied}, research on fairness can be categorized into three aspects: pre-processing methods, in-processing methods, and post-processing methods \cite{mehrabi2021survey, dai2022comprehensive}.
Pre-processing methods endeavor to transform the training data in a manner that eliminates underlying discrimination before model training~\cite{d2017conscientious, kang2020inform}.
In-processing methods are designed to incorporate fairness considerations into the training stage of SOTA models to mitigate discrimination during the training process~\cite{dai2020learning, bose2019compositional}.
Post-processing methods directly modify the prediction results generated by a given model to ensure fairness.

In this paper, we introduce an in-processing FL framework specifically designed to ensure multi-group fairness within the non-IID FL scenario.

\subsection{Fair Federated Learning}
Fairness remains an active topic within the realm of FL research. 
The existing literature on fairness in FL predominantly concentrated on particular fairness notions introduced in FL, including client-based fairness \cite{mammen2021federated, li2021ditto} and collaborative fairness \cite{mammen2021federated, lyu2020collaborative}. 
%
However, the impact of FL on group fairness has not been comprehensively understood to date.

Recently, considerable progress has been made in training models with group fairness guarantees in the context of FL.
Based on different fairness requirements, research can be generally divided into two different categories: local fairness and global fairness \cite{ezzeldin2023fairfed, zeng2021improving}.

\textit{For local fairness}, the goal is to find a model that satisfies each fairness requirement in each local model \cite{ zeng2021improving}. 
Some researchers \cite{chang2023bias} provided empirical evidence that engagement in FL can potentially have a detrimental effect on group fairness.
Some studies \cite{cui2021addressing, papadaki2022minimax} proposed algorithms to enhance local fairness for clients without sacrificing performance consistency. 

\textit{For global fairness}, it aims to achieve a single fairness requirement on the global data distribution across all participating clients.
In this paper, we concentrate on global fairness in FL, which can be divided into two categories.
(1) Reweighting techniques \cite{zeng2021improving, ezzeldin2023fairfed, mohri2019agnostic, papadaki2022minimax}, which dynamically reweights clients or data during the training process.
The main purpose of dynamical reweighting techniques is to equalize the learning loss on each sensitive group or fairness loss on each client. 
This approach is motivated by FairBatch \cite{roh2020fairbatch}, which demonstrates that maintaining consistent 0-1 loss across all groups serves as the sufficient condition for achieving group fairness. 
Some researchers~\cite{zeng2021improving} adapted the FairBatch multi-group debiasing algorithm into FL.
(2) Distributively solve an optimization objective with fairness constraints or fairness regularization \cite{ezzeldin2023fairfed, du2021fairness, wang2023mitigating, dunda2023handling, zhang2020fairfl}.
The common approach for handling the non-convex and non-differentiable fairness constraints in these works is to utilize the surrogate function to approximate the real classification result \cite{du2021fairness, wang2023mitigating, dunda2023handling, zhang2020fairfl}.
Furthermore, most optimization methods are tailored for specific two-group fairness measures, lacking the ability to address multi-group situations and other fairness measures, thus demonstrating limited scalability and flexibility in intricate FL scenarios \cite{kairouz2021advances}.
Nevertheless, within the FL setting, data heterogeneity can detrimentally impact model performance, primarily because of the limitations inherent in surrogate functions, as illustrated in Figure~\ref{introduction-global-unfair}.
In our work, the proposed WassFFed method can achieve multi-group fairness in case of data heterogeneity by enforcing the distribution of sensitive groups’ classification toward the Wasserstein barycenter. 
Moreover, the \M~method directly manipulates the output scores of the classification model, eliminating the necessity for surrogate function computations.

\section{Problem Formulation}
This paper focuses on achieving group fairness within the context of FL.
The problem formulation can be divided into two key components: federated learning and group fairness.

\subsection{Federated Learning}
FL addresses scenarios in which users' data cannot be shared across clients due to privacy concerns.
Consequently, FL enables clients to train their local models using their respective local datasets and then aggregate these local models to construct a global model.
Let $\mathcal{D} = \{\mathcal{X}, \mathcal{Y}\}$ represents the global data distribution, where $\mathcal{X}$ denotes the input space, and $\mathcal{Y}$ denotes the output space.
In this paper, we consider the binary classification task that $\mathcal{Y} = \{0, 1\}$. 
Each client $C_p$ possesses access to its private local dataset, denoted as $\mathcal{D}_p = \{\mathcal{X}_p, \mathcal{Y}_p\}$.
Each sample in $\mathcal{D}_p$ is represented as $t_i^p: (x_i^p, y_i^p)$, where $i \in [1, N_p]$, and $N_p$ represents the number of samples in the local dataset of client $C_p$.
Consequently, the total data distribution is given by $\mathcal{D} = \cup_{p \in [1, P]} \mathcal{D}_p$.

In FL, each client trains a local model represented as $\hat{y}_i^p = f_p (x_i^p; w_p)$, with $w_p$ denoting the parameter set of $f_p$.
Then, the server aggregates these local models to construct a global model denoted as $f$.
The overall objective in FL is defined as follows:
\begin{equation}
\begin{split}
    \label{FL-objective}
    & \min L(w_1, w_2, \dots, w_P; \mathbf{\lambda}) \\ &= \sum_{p = 1}^P \lambda_p \mathbb{E}_{(x_i^p, y_i^p) \sim \mathcal{D}_p}[L_p(f_p (x_i^p;w_p), y_i^p)],
\end{split}
\end{equation}
where $L_p$ represents the local loss of client $p$, and $\lambda_p$ signifies the weight ratio used for aggregation.

\subsection{Group Fairness}
The concept of group fairness aims to ensure that ML models provide equitable treatment to users with diverse sensitive attributes, such as gender, race, and age.
Let $\mathcal{A} = \{a_1, a_2, \dots, a_{N_A}\}$ represent the set of sensitive groups, where each group corresponds to users sharing a specific value of a kind of sensitive attribute, and $N_A$ denotes the total number of such groups.
There are two major categories of group fairness quantification, we give the definitions as follows:
\begin{definition}[Demographic Parity (DP)~\cite{dwork2012fairness}]
\begin{equation}
\label{demographic-parity-difinition}
    P(\hat{Y} = 1 | A = a_1) = P(\hat{Y} = 1 | A = a_2) = \dots = P(\hat{Y} = 1 | A = a_{N_A}).
\end{equation}
\end{definition}
\begin{definition}[Equal Opportunity (EOP)~\cite{hardt2016equality}]
\begin{equation}
\label{eop-difinition}
    P(\hat{Y} = 1 | A = a_1, Y = 1) = \dots = P(\hat{Y} = 1 | A = a_{N_A}, Y = 1).
\end{equation}
\end{definition}
DP focuses on achieving an equal positive prediction rate among different groups, while EOP concentrates on attaining the same true positive rate across those groups.
To ensure comprehensive fairness in prediction results, we require the FL framework to satisfy both DP and EOP.
However, DP and EOP do not directly provide a fairness measurement for prediction models.
Therefore, we define two metrics to assess the fairness level of machine learning models:
\begin{definition}[Metric of Demographic Parity ($\mathcal{M}_{DP}$)]
\begin{equation}
\label{demographic-parity-metric}
\begin{split}
    \mathcal{M}_{DP} = max\{|\mathbb{E}[\hat{Y} = 1 | A = a_i] - \mathbb{E}[\hat{Y} = 1 | A = a_j]|\}, 
    \\ \forall a_i, a_j \in \mathcal{A}, a_i \ne a_j.
\end{split}
\end{equation}
\end{definition}
\begin{definition}[Metric of Equal Opportunity ($\mathcal{M}_{EOP}$)]
\begin{equation}
\label{equal-opportunity-metric}
\begin{split}
    \mathcal{M}_{EOP} = max\{|\mathbb{E}[\hat{Y} = 1 | A = a_i, Y = 1] \\ - \mathbb{E}[\hat{Y} = 1 | A = a_j, Y = 1]|\},
    \forall a_i, a_j \in \mathcal{A}, a_i \ne a_j.
\end{split}
\end{equation}
\end{definition}

Obviously, smaller values of $\mathcal{M}_{DP}$ and $\mathcal{M}_{EOP}$ indicate fairer models.
In this paper, our goal is to strike a balance between accuracy and fairness within the context of FL.

\section{Methodology}
In this section, we provide a comprehensive introduction to the \M~framework.
We will begin with a brief overview of \M, followed by an in-depth exploration of each modeling stage.
\begin{figure*}[htbp]
  \centering
  \includegraphics[width=\linewidth]{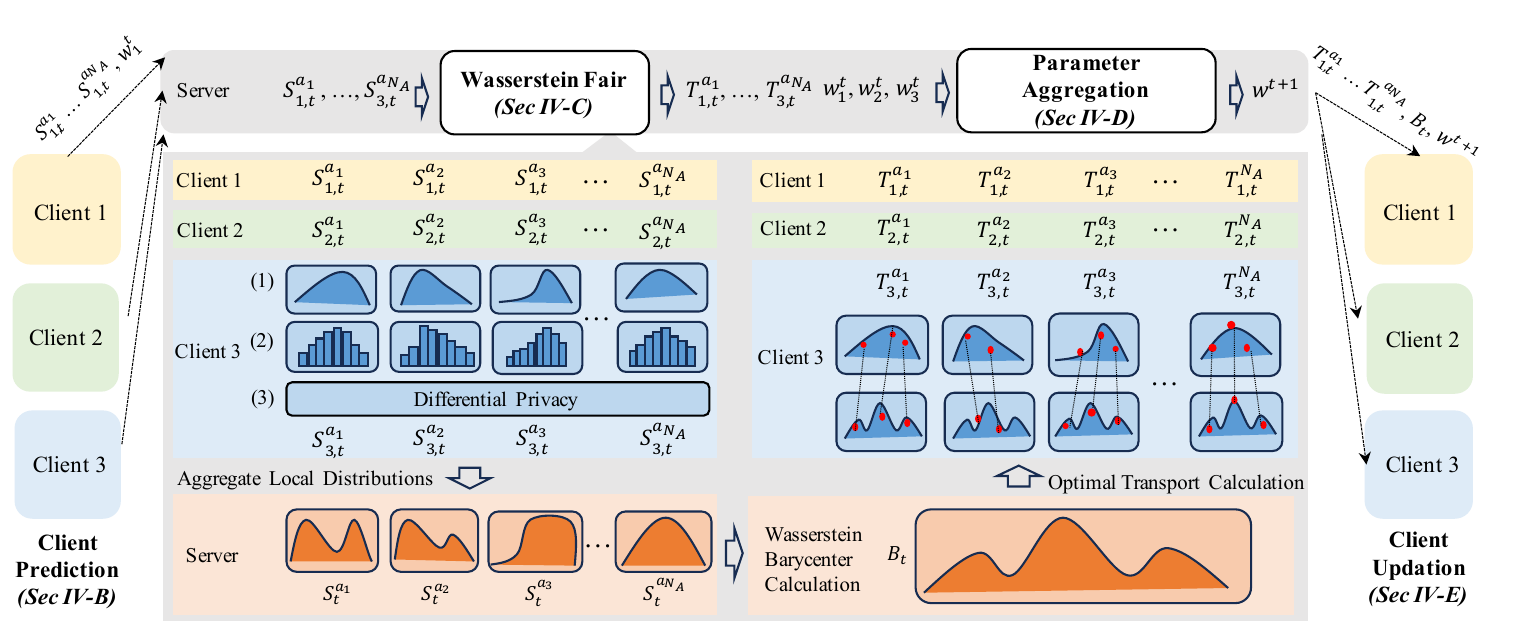}
      \caption{The overall framework of \M. We give an example of three clients. 
      Firstly, in the Client Prediction stage, all clients share their parameters ($w_t$) and model outputs for various sensitive groups ($S_{1, t}^{a_1}, S_{1, t}^{a_2}, \dots, S_{3, t}^{a_{N_A}}$) with the server. Subsequently, the server employs the Wasserstein Fair model to compute transport matrices ($T_{1, t}^{a_1}, T_{1, t}^{a_2}, \dots, T_{3, t}^{a_{N_A}}$) and aggregates parameters ($w^{t+1}$). Finally, in the Client Updation stage, the server shares these results with each client. Clients calculate the fairness loss and combine it with the model utility loss to strike a balance between accuracy and fairness.
      }
      \label{overall-framework}
\end{figure*}

\subsection{Overview}
In this paper, we propose a novel \textbf{Wass}erstein \textbf{F}air \textbf{Fed}erated Learning framework, namely \M, to strike a balance between accuracy and fairness in the context of FL.
As shown in Figure~\ref{overall-framework}, the overall framework of \M~comprises four stages.
(1) In the \textit{Client Prediction} stage, all clients share their model parameters and encrypted model outputs for various sensitive groups with the server.
(2) In the \textit{Wasserstein Fair} stage, which constitutes the primary contribution of this paper, the server aggregates all output distributions from clients and calculates a Wasserstein barycenter.
Subsequently, for each client, the server computes the optimal transport matrices for each output distribution corresponding to a sensitive group.
The server's objective is to bring all clients' outputs corresponding to users with different sensitive attributes closer to the barycenter, ensuring that model prediction results are independent of sensitive attributes.
(3) In the \textit{Parameter Aggregation} stage, the server aggregates all clients' parameters to update models.
(4) In the \textit{Client Updation} stage, all clients calculate a fairness loss based on received transport matrices.
\M~combines the fairness loss with the model utility loss to optimize local models in both a fair and accurate direction.

\subsection{Client Prediction}
In the beginning, all clients train their local models and generate model outputs from their local datasets.
In each client, users with different sensitive attributes are divided into different sensitive groups, i.e., $a_1, a_2, \dots, a_{N_A}$.
Taking client 3 in Figure~\ref{overall-framework} as an example, \M~aggregates all model outputs for each group, i.e., $S_{3,t}^{a_1}, S_{3,t}^{a_2}, \dots, S_{3,t}^{a_{N_A}}$, during the $t$-th communication round with the server, resulting in overall output distributions for each group.
We outline three steps to process model outputs, enhancing the efficiency of subsequent computations and safeguarding client privacy.
These steps are elaborated in Section~\ref{computation-method-section}.

The inherent challenge of group fairness~\cite{du2021fairness,zafar2017parity} always leads to disparate treatment of users with different sensitive attributes by prediction models.
Consequently, $S_{3,t}^{a_1}, S_{3,t}^{a_2}, \dots, S_{3,t}^{a_{N_A}}$ may exhibit notably different distributions, thereby yielding unfair prediction outcomes.
\M~addresses this issue by striving to mitigate the discrepancies among all clients.
It achieves this by enforcing the similarity of output distributions for various sensitive groups to a certain global distribution.
As a result, model outputs become independent of sensitive attributes.
It's important to note that \M~primarily focuses on the model output values, rather than the classification results involving thresholds or model loss calculated using surrogate functions.
This approach enables \M~to circumvent the essential estimation errors associated with the use of surrogate functions, ensuring fair prediction outcomes across various thresholds.
During this stage, all clients share their model output distributions and model parameters ($w_1^t, w_2^t, \dots, w_P^t$) with the server.

\subsection{Wasserstein Fair}
In this stage, the server aggregates the output distributions from all clients and calculates the corresponding Wasserstein barycenter.
This barycenter is treated as a global distribution, and the primary objective of \M~is to ensure that the output distributions of various sensitive groups from all clients closely resemble this global distribution.
Consequently, the server calculates optimal transport matrices to guide the distributions of each client towards the barycenter.
This process ensures that \M~avoids any inconsistency between the global model and local models, as all clients share the same global distribution.

\nosection{Aggregate Local Distributions}
After receiving output distributions from clients, the server needs to first construct global output distributions for various sensitive groups.
This step is particularly crucial as the data distribution in each client may exhibit non-IID characteristics, potentially resulting in certain sensitive groups being rare in specific clients.
The global distribution aggregation can be calculated as follows:
\begin{equation}
    \label{global-distribution-aggregation}
    S_t^{a_i} = \cup_{p \in [1, P]} S_{p, t}^{a_i}, \quad \forall a_i 
    \in \mathcal{A},
\end{equation}
where $S_t^{a_i}$ denotes the global output distribution for sensitive group $a_i$ in communication round $t$.

\nosection{Wasserstein Barycenter Calculation}
In this step, the server calculates a uniform global distribution that aggregates characteristics from the output distribution for each sensitive group.
Wasserstein barycenter\cite{cuturi2014fast} introduces an approach to compute a central distribution for several distributions while minimizing the distance between these distributions and the barycenter.
This approach is particularly suitable for our task since we aim to align the output distribution of each sensitive group with the global distribution.
The minimized distance achieved through the Wasserstein barycenter method ensures the efficiency of this alignment.

To calculate the Wasserstein barycenter, firstly, we provide the definition of the $q$-Wasserstein distance.
Wasserstein distance assesses the minimum cost of transporting a distribution to another, i.e., solving the Optimal Transport (OT)~\cite{villani2009optimal} problem.
Given two distributions $S_1$ and $S_2$, let $\mathcal{T}: \{S_1 \times S_2 \rightarrow [0, +\infty]\}$ be the set of transport maps from $S_1$ to $S_2$, and $C: S_1 \times S_2 \rightarrow [0, +\infty]$ be the cost function such that $C(s_1, s_2)$ indicates the cost of transporting $s_1$ to $s_2$.
Then, the optimal transport problem~\cite{bogachev2012monge} is formulated as:
\begin{equation}
    \label{optimal-transport-definition}
    T^* = \underset{T \in \mathcal{T}}{\arg\min} \int_{S_1 \times S_2} C(s_1, s_2) T(s_1, s_2)d s_1 d s_2.
\end{equation}
$T^*$ denotes the optimal transport matrix that minimizes the total transport cost.
Based on the idea of OT, $q$-th Wasserstein distance is defined as:
\begin{equation}
    \label{q-th-wasserstein-distance}
    \mathcal{W}_q(S_1, S_2) = \min_{T \in \mathcal{T}} \left(\int_{S_1 \times S_2} C(s_1, s_2) ^ q T(s_1, s_2) ds_1 ds_2 \right)^{\frac{1}{q}}.
\end{equation}
In this paper, we focus on the distribution of model outputs, which is always one-dimensional.
Therefore, we employ the 1-Wasserstein distance~\cite{ruschendorf1985wasserstein} to quantify the divergence between two distributions:
\begin{equation}
\begin{split}
    \label{1-th-wasserstein-distance}
    \mathcal{W}_1(S_1, S_2) &= \min_{T \in \mathcal{T}} \int_{S_1 \times S_2} C(s_1, s_2) T(s_1, s_2) ds_1 ds_2
    \\ &= \int_{S_1 \times S_2} \left|s_1 - s_2\right| T^*(s_1, s_2) ds_1 ds_2.
\end{split}
\end{equation}
Subsequently, we utilize this metric to calculate the Wasserstein Barycenter $B_t$ as follows:
\begin{equation}
    \label{wasserstein-barycenter-calculation}
    B_t = \underset{B \in \mathcal{B}}{\arg\min} \sum_{a \in \mathcal{A}}\lambda_t^a \mathcal{W}_1(B, S_t^a),
\end{equation}
where $\mathcal{B}$ denotes the set of essential barycenters and $\lambda_t^a$ indicates the weight ratio used for aggregation.

\nosection{Optimal Transport Calculation}
\M~treats $B_t$ as the global output distribution and aims to enforce all clients' output distributions to be closer to that.
To achieve this goal, it requires minimal changes in model predictions to preserve prediction accuracy as much as possible.
For two distribution $S_1$ and $S_2$, the triangle inequality, $\mathcal{W}_1(S_1, B_t) \leq \left|\mathcal{W}_1(S_1, S_2) + \mathcal{W}_1(S_2, B_t)\right|$~\cite{jiang2020wasserstein}, proves that the distance $\mathcal{W}_1(S_1, B_t)$ reaches minimum if and only if $S_2$ lies on the shortest path between $S_1$ and $B_t$.

The above discussion illustrates that by training each client model along the optimal transport path using the gradient descent method, each client can produce fair results with minimal adjustments.
Therefore, for a sensitive group $a$ in client $p$, the server calculates the optimal transport matrix based on $1$-th Wasserstein distance as follows:
\begin{equation}
    \label{optimal-transport-calculating}
    T_{p, t}^{a} = \underset{T \in \mathcal{T}}{\arg\min} \int_{S_{p, t}^a \times B_t} \mathcal{W}_1(s_{p, t}^a, b_t) T(s_{p, t}^a, b_t)d s_{p, t}^a db_t.
\end{equation}
However, calculating this optimal transport problem may be time-consuming, with a worst-case time complexity of $O(N^3\log N)$, where $N$ is the dimension of $S_{p, t}^a$.
To overcome this, we introduce the sinkhorn divergence~\cite{cuturi2013sinkhorn} to smooth the objective:
\begin{equation}
\begin{split}
    \label{sinkhorn-calculation}
    T_{p, t}^{a} &= \underset{T \in \mathcal{T}}{\arg\min} \int_{S_{p, t}^a \times B_t} \mathcal{W}_1(s_{p, t}^a, b_t) T(s_{p, t}^a, b_t)d s_{p, t}^a db_t
    \\ &+ \epsilon \cdot \int_{S_{p, t}^a \times B_t}T(s_{p, t}^a, b_t)(\log(T(s_{p, t}^a, b_t)) - 1) d s_{p, t}^a db_t,
\end{split}
\end{equation}
where $\epsilon$ controls the importance of the entropy.
We introduce the detailed optimization process for~\eqref{sinkhorn-calculation} in Section~\ref{sec: computation}

In this step, the server calculates transport matrices of various sensitive groups for all clients and transfers the results to each client, together with the barycenter.
For instance, client 1 in Figure~\ref{overall-framework} receives $T_{1, t}^{a_1}, T_{1, t}^{a_2}, \dots, T_{1, t}^{a_{N_A}}$ and $B_t$.

\subsection{Parameter Aggregation}
In addition to computing the optimal transport paths, the server also aggregates the parameters from clients.
In this stage, \M~employs the FedAvg method to aggregate parameters $w_1^t, w_2^t, \dots, w_P^t$ and transfers the model parameters for the next round, denoted as $w_{t+1}$, to each client.

\subsection{Client Updation}
After receiving transport matrices and model parameters from the server, each client updates its parameters and trains its local model for $k$ rounds.
Taking client 1 as an example, in each round, it first calculates the output distributions of various sensitive attributes, denoted as $S_1^{a_1}, S_1^{a_2}, \dots, S_1^{a_{N_A}}$.
Then, client 1 computes the fairness loss following the $1$-Wasserstein distance:
\begin{equation}
\begin{split}
    \label{fairness-loss}
    L_{fairness} &= \sum_{a \in \mathcal{A}} \sum_{s \in S_1^a} \sum_{b \in B_t} \mathcal{W}_1(S_1^a, B_t)
    \\ &= \sum_{a \in \mathcal{A}} \sum_{s \in S_1^a} \sum_{b \in B_t} |s - b| T_{1, t}^a(s, b).
\end{split}
\end{equation}
All clients combine the fairness loss with the original model loss, denoted as $L_{utility}$ to compute the final loss function:
\begin{equation}
    \label{final-loss}
    L = \beta L_{utility} + (1 - \beta) L_{fairness},
\end{equation}
where $\beta$ controls the trade-off between accuracy and fairness.
By minimizing this loss, the model can achieve a balance between accuracy and fairness.

Note that in the first $k$ rounds, clients only compute $L_{utility}$ to initialize local models.
We summarize the overall process of \M~in Algorithm~\ref{algorithm-total}.
Note that in the first $k$ rounds, each client only computes the model utility loss $L_{utility}$ to initialize the local model.
\begin{algorithm}[!h]
\caption{\M}\label{algorithm-total}
    \SetKwData{Left}{left}\SetKwData{This}{this}\SetKwData{Up}{up}
    \SetKwFunction{Union}{Union}\SetKwFunction{FindCompress}{FindCompress}
    \SetKwInOut{Input}{Input}\SetKwInOut{Output}{Output}
    \Input{Datasets $\mathcal{D}_p$ from client $p$, $p = 1, 2, \dots, P$; Training steps $\tau$;  Initial parameters $w^0$; Balance hyperparameter $\beta$; local training round $k$;}
    \Output{Final parameters $w$;}
    $t = 0$; \\
    \textbf{Client Side:}\\
    Initialize local models with $w^0$; \\
    \For{$i = 1, 2, \dots, k$}{
        Train local models with $L_{utility}$;
    }
    \For{$j = 1, 2, \dots, P$}{
        Client $j$ calculates $S_{j, t}^{a_1}, S_{j, t}^{a_2}, \dots, S_{j, t}^{a_{N_A}}$ and transfer them to the server together with parameters $w_j^t$;
    }
    \While{$t < \tau$}{
    \textbf{Server Side:}\\
    Calculate global distributions $S_t^{a_1}, S_t^{a_2}, \dots, S_t^{a_{N_A}}$;\\
    Calculate the Wasserstein barycenter $B_t$; \\
    Calculate $w^{t+1}$;\\
    \For{$j = 1, 2, \dots, P$}{
        Calculate optimal transport matrices  $T_{j, t}^{a_1}, T_{j, t}^{a_2}, \dots, T_{j, t}^{a_{N_A}}$ and transfer to client $j$ together with $B_t$, and $w^{t+1}$; \\ 
    }
    
    \textbf{Client Side:}\\
    \For{$i = 1, 2, \dots, k$}{
    \For{$j = 1, 2, \dots, P$}{
    Update model parameters with $w^{t+1}$;\\
        Calculate distributions $S_j^{a_1}, S_j^{a_2}, \dots, S_j^{a_{N_A}}$; \\
        Calculate loss $L = \beta L_{utility} + (1-\beta)L_{fairness}$ and update the model; \\
    }
    }
    $t++$\\
    }
    \textbf{return} final parameters $w$ aggregated by the server;
    \label{algorithm}
\end{algorithm}

\section{Computation and Analysis of \M \label{sec: computation}}
In this section, we describe the computation method of \M, and provide an in-depth analysis of its efficiency.

\subsection{Computation Method \label{computation-method-section}}
We detail the computation method of \M, which prioritizes both computational efficiency and privacy preservation.
The computation method comprises two main components: the calculation of model outputs and the optimal transport matrices.
As illustrated in Figure~\ref{overall-framework}, we take client $3$ as an example. 
 
\nosection{Calculation of Model Outputs}
The calculation of model outputs $S_{3,t}^{a_1}, S_{3,t}^{a_2}, \dots, S_{3,t}^{a_{N_A}}$ includes three steps.
(1) Initially, \M~generates the original model output distribution for each sensitive group within every client.
(2) Consequently, \M~employs a widely recognized strategy that assigns the support of distributions to uniformly distributed bins across the $[0, 1]$ interval.
For example, an output of $0.4325$ from a client's model, with the bin number ($N_B$) set to $10$, would be allocated to the bin $[0.4, 0.5)$. 
Barycenters are computed similarly.
(3) Finally, \M~employs the differential privacy~\cite{dwork2014differential_privacy} approach to further protect user privacy, by applying randomized responses on the histogram of the output.
Specifically, for each output value, with a small probability $\xi$ we allocate it to a uniformly random bin, directly leading to a $(\ln \xi)$-differential privacy for the user output.

The above approach offers dual benefits, \textit{on the one side}, it enables the use of the iterative KL-projection method for efficient barycenter approximation, confirmed to be time-efficient with a complexity of $\mathcal{O}(M\log M)$ for a barycenter comprising $M$ samples.
\textit{On the other side}, It safeguards user privacy by approximating outputs as coarse histograms and utilizing the differential privacy approach.


\nosection{Calculation of Optimal Transport Matrices}
To calculate the optimal transport matrices $T_{1, t}^{a_1}, \dots, T_{P, t}^{a_{N_A}}$, we need to solve the Equation~\eqref{sinkhorn-calculation}, 
we provide the details of optimizing this objective with the sinkhorn divergences follows.
Since in practice, we always calculate the optimal transport problem based on data points instead of a distribution, we rewrite the object in a discrete form:
\begin{equation}
\begin{split}
\label{sinkhorn-algorithm-discrete}
     T_{p, t}^{a} &= \underset{T \in \mathcal{T}}{\arg\min} \sum_{s \in S_{p, t}^a}\sum_{b \in B_t} C(s,b) T(s,b) 
     \\ &+ \epsilon \cdot \sum_{s \in S_{p, t}^a}\sum_{b \in B_t} T(s,b)(\log(T(s,b)) - 1).
\end{split}
\end{equation}
We rewrite (\ref{sinkhorn-algorithm-discrete}) with Lagrange multipliers as
\begin{align}\label{equ:lag}
    \max_{\boldsymbol{f}, \boldsymbol{g}}\min_{\boldsymbol{T}} \mathcal{J} = \Bigg\{ & \sum_{s \in S_{p, t}^a} \sum_{b \in B_t} C(s,b)  T(s,b) \nonumber \\
        &+ \epsilon \cdot \sum_{s \in S_{p, t}^a} \sum_{b \in B_t} T(s,b)(\log(T(s,b)) - 1) \nonumber \\
        &- \sum_{b \in B_t} f_b \Bigg[\Bigg(\sum_{s \in S_{p, t}^a} T(s,b)\Bigg)
        - \frac{1}{|B_t|}\Bigg] \nonumber\\
    & - \sum_{s \in S_{p, t}^a} g_s \Bigg[\Bigg(\sum_{b \in B_t} T(s,b)\Bigg) - \frac{1}{|S_{p, t}^a|}\Bigg]\Bigg\}.
\end{align}
Taking the differentiation w.r.t. $T(s,b)$ on (\ref{equ:lag}), we have
\begin{equation}
    \frac{\partial\mathcal{J}}{\partial T(s,b)} = 0 \enspace \Rightarrow \enspace C(s,b) + \epsilon \cdot \log(T(s,b)) - f_b - g_s = 0.
\end{equation}
To update our variables, we first fix $g_s$ and update $f_b$ with
\begin{equation}
    f_b^{(t+1)} = \epsilon \cdot \Bigg\{\log\bigg(\frac{1}{|B_t|}\bigg) - \log\Bigg[\sum_{s \in S_{p, t}^a} \exp\bigg(\frac{g_s^{(t)} - C(s,b))}{\epsilon}\bigg)\Bigg]\Bigg\}.
\end{equation}
Then we fix $f_b$ and update $g_s$ with
\begin{equation}
    g_s^{(t+1)} = \epsilon \cdot \Bigg\{\log\bigg(\frac{1}{|S_{p, t}^a|}\bigg) - \log\Bigg[\sum_{b \in B_t} \exp\bigg(\frac{f_b^{(t)} - C(s,b))}{\epsilon}\bigg)\Bigg]\Bigg\}.
\end{equation}
In summary, we can iteratively update $f_b$ and $g_s$ until we obtain the final solutions.
Considering that the output distributions of models are consistently one-dimensional, the above calculation method can be simplified~\cite{deshpande2018generative, jiang2020wasserstein} with a time complexity of $\mathcal{O}(M\log M)$.

\subsection{Analyze the Efficiency of \M}
This section examines the efficiency of the proposed \M~framework, focusing on computational efficiency, communication efficiency, and privacy considerations.

\nosection{Computation Efficiency}
As outlined in Section~\ref{computation-method-section}, the incremental time cost associated with \M~is $\mathcal{O}(M\log M)$, indicating a time-efficient approach.

\nosection{Commuication Efficiency}
The \M~framework's additional communication between the server and clients primarily involves transmitting client model outputs, optimal transport matrices, and the Wasserstein barycenter.
\textit{For client model outputs,} it is sufficient to transfer the value of each bin to the server, denoted as $N_B$ numbers. 
\textit{Regarding optimal transport matrices}, literature such as ~\cite{cuturi2013sinkhorn, jiang2020wasserstein} demonstrates that an optimal transport matrix between distributions $S_1$ and $S_2$ can have at most $\mathcal{O}(|S_1| + |S_2|)$ nonzero entries, which are needed to transfer between the server and clients.
\textit{For the Wasserstein barycenter}, we only need to transfer the value of each bin to each client, denoted as $M$ numbers. 
Thus, the extra communication costs brought by \M~is limited with space complexity of $\mathcal{O}(N)$.

\nosection{Privacy}
To safeguard client privacy, two methods are employed: organizing model outputs into coarse bins and applying differential privacy, as discussed in Section~\ref{computation-method-section}.
These techniques introduce a level of approximation and randomness to client outputs, thereby enhancing user privacy.

\section{Experiments and analysis}

\begin{table}\centering 
\renewcommand{\arraystretch}{1.1}
\caption{The statistics of datasets}
\label{dataset-statistic}
\begin{threeparttable}
\resizebox{\linewidth}{!}{
\begin{tabular}{llll}
\hline
Dataset & Samples & Model & Sensitive attributes \\ \hline
Adult & 46,447 & Logistic regression &  \multirow{3}{*}{\shortstack{\textbf{Race}: white, non-white; \\ \textbf{Gender}: male, female.}} \\ \cline{1-3}
Compas & 6,819 & Logistic regression & \\ \cline{1-3}
CelebA & 202,599 & ResNet18 & \\
\hline
\end{tabular}
}
\vspace{-10pt}
    \end{threeparttable}
\end{table}


%
To comprehensively assess the proposed \M~framework, we conduct extensive experiments on three publicly available real-world datasets to answer the following Research Questions (RQ): \textbf{RQ1}: Does \M~outperform existing methods in effectively achieving a balance between accuracy and fairness?
\textbf{RQ2}: What is the impact of calculating a global Wasserstein barycenter on the enhancement of model performance?
\textbf{RQ3}: Can \M~ensure that the model outputs for different sensitive groups become more similar?
\textbf{RQ4}: How do important hyperparameters influence the performance of \M?
\textbf{RQ5}: How does the number of clients influence the performance of \M?


\subsection{Datsets and Experimental Settings}
In this section, we present the experimental setup of the paper, covering the datasets, baselines, evaluation protocols, and parameter settings.

\nosection{Datasets}
We conduct an evaluation of our proposed \M~on three publicly available real-world datasets, Adult~\cite{asuncion2007uci}, Compas~\cite{dressel2018accuracy}, and CelebA~\cite{zhang2020celeba}.
These datasets are well-established for assessing fairness issues in FL~\cite{zeng2021improving,du2021fairness,wang2023mitigating,abay2020mitigating}.
We summarize the statistics of these datasets in Table~\ref{dataset-statistic}.
For the \textbf{Adult} dataset, the task involves predicting whether an individual's annual income exceeds $50,000$ or not.
In the case of the \textbf{Compas} dataset, the prediction centers around whether individuals who have previously committed legal infractions within the past two years will re-offend.
Lastly, the \textbf{CelebA} dataset entails predicting whether the individuals in the images exhibit a smiling expression.
In all these datasets, we have identified \textit{race} and \textit{gender} as sensitive attributes, following the methodology in~\cite{jiang2020wasserstein, jiang2020identifying}.
For the relatively smaller datasets, Adult and Compas, we employed logistic regression~\cite{lavalley2008logistic} for training on both clients and the server.
However, for the larger CelebA dataset, which involves a more complex prediction task, we opted for the ResNet18 model~\cite{he2016deep}.
This approach enables us to comprehensively evaluate the performance of \M~in practical scenarios, spanning different data scales and a variety of tasks.

\begin{table*}\centering 
\renewcommand{\arraystretch}{1.1}
\caption{Experimental result with multi-sensitive group}
\label{multi-sensitive-experiment-result-table}
\begin{threeparttable}
\resizebox{\linewidth}{!}{
\begin{tabular}{clccccccccc}
\hline
& Dataset & \multicolumn{3}{c}{Adult} & \multicolumn{3}{c}{Compas} & \multicolumn{3}{c}{CelebA}  \\
\cmidrule(lr){3-5}\cmidrule(lr){6-8}\cmidrule(lr){9-11}

$\alpha$ & Method & Acc ($\uparrow$) & $\mathcal{M}_{DP}$ ($\downarrow$) & $\mathcal{M}_{EOP}$ ($\downarrow$) & Acc ($\uparrow$) & $\mathcal{M}_{DP}$ ($\downarrow$) & $\mathcal{M}_{EOP}$ ($\downarrow$)  & Acc ($\uparrow$) & $\mathcal{M}_{DP}$ ($\downarrow$) & $\mathcal{M}_{EOP}$ ($\downarrow$) \\ \hline

\multirow{6}{*}{0.1} & FedAvg & 0.8419* & 0.2126 & 0.1732 & 0.6860* & 0.3014 & 0.2832 & 0.8949* & 0.2513 & 0.1040 \\ 

& AFL & 0.8042 & \underline{0.0989} & {0.1416} & 0.6478 & \underline{0.2173} & 0.2274 & 0.8463 & 0.1878  & {0.0978} \\

& LocalFair & 0.8127 & 0.1273 & 0.1599 & 0.6093 & 0.2642 & 0.2592 & 0.8651 & 0.2165 & 0.0819   \\

& FEDFB & 0.8119 & 0.1039 & \underline{0.1250} & 0.6228 & 0.2178 & \underline{0.1810} & \underline{0.8823} & 0.1991 & \underline{0.0733}  \\

& \MLocal & 0.8058 & 0.1013 & 0.1402 & \underline{0.6493} & 0.2203 & 0.1995 & 0.8792 & \underline{0.1965} & 0.1425 \\ 

& \M & \textbf{\underline{0.8141}} & \textbf{0.0937}* & {\textbf{0.1198}*} & \textbf{0.6336} & \textbf{0.1243}* & \textbf{0.1623}* & \textbf{0.8708} & \textbf{{0.1312*}} & \textbf{0.0571}* \\ \hline

\multirow{6}{*}{0.5} & FedAvg & 0.8422* & 0.2161 & 0.1703 & 0.6879* & 0.3048 & 0.2849 & 0.8978* & 0.2619 & 0.1104 \\ 

& AFL & 0.8122 & {0.1523} & {0.1519} & 0.6346 & \underline{0.1911} & \underline{0.1561} & 0.8247 & 0.2032  & {0.0732*} \\

& LocalFair & 0.8133 & 0.1117 & \underline{0.1310} & 0.6435 & 0.2638 & 0.2360 & 0.8752 & 0.2307 & 0.0999   \\

& FEDFB & 0.8121 & \underline{0.1096} & 0.1539 & 0.6315 & 0.2576 & 0.2357 & {0.8784} & \underline{0.1727} & 0.0846  \\

& \MLocal & 0.8125 & 0.1106 & 0.1541 & \underline{0.6625} & 0.2171 & 0.2298 & \underline{0.8792} & {0.1942} & 0.0873 \\ 

& \M & \textbf{\underline{0.8153}} & \textbf{0.0911}* & \textbf{0.1186}* & \textbf{0.6355} & \textbf{0.1272}* & \textbf{0.1533}* & \textbf{0.8755} & \textbf{{0.1421*}} & \textbf{\underline{0.0751}} \\ \hline

\multirow{6}{*}{5} & FedAvg & 0.8423* & 0.2183 & 0.1745 & {0.6884}* & 0.3041 & 0.2915 & 0.9037* & 0.2626 & 0.1280  \\

& AFL & 0.8045 & 0.1278 & 0.1330 & 0.6375 & 0.2310 & 0.2541 & 0.8135 & \underline{0.1662} & 0.0803   \\

& LocalFair & 0.8129 & 0.1313 & 0.1685 & 0.6451 & 0.2644 & {0.2501} & 0.8805 & {0.1903} & {0.0970}  \\

& FEDFB & \underline{0.8173} & 0.1261 & 0.1423 & 0.6460 & 0.2735 & 0.2634 & 0.8846 & 0.1846 & 0.0798  \\

& \MLocal & 0.8120 & \underline{0.0984} & \underline{0.1357} & \underline{0.6661} & \underline{0.2173} & \underline{0.2240} & 0.8900 & 0.1694  & \underline{0.0756} \\

& \M & \textbf{0.8152} & \textbf{{0.0897*}} & \textbf{{0.1216*}} & \textbf{0.6451} & \textbf{0.1425}* & \textbf{0.1712}* & \textbf{\underline{0.8852}} & \textbf{0.1614}* & \textbf{0.0702}* \\ \hline

\multirow{6}{*}{20} & FedAvg & 0.8427* & 0.2108 & 0.1657 & {0.6889}* & 0.3067 & {0.2948} & 0.9090* & 0.2714 & 0.1309  \\

& AFL & 0.8095 & 0.1126 & 0.1317 & 0.6375 & 0.2032 & 0.2267 & 0.8598 & 0.1769 & 0.0908   \\

& LocalFair & 0.8178 & 0.1542 & 0.1539 & 0.6324 & 0.2714 & {0.2683} & 0.8734 & {0.2092} & {0.1175}  \\

& FEDFB & {0.8153} & {0.1141} & \underline{0.1309} & 0.6413 & \underline{0.1597} & 0.2146 & 0.8825 & \underline{0.1757} & 0.0824  \\

& \MLocal & 0.8097 & \underline{0.1098} & {0.1566} & \underline{0.6490} & 6{0.1854} & \underline{0.1907} & 0.8891 & 0.1819  & \underline{0.0840} \\

& \M & \underline{\textbf{0.8177}} & \textbf{{0.0918*}} & \textbf{{0.1263*}} & \textbf{0.6399} & \textbf{0.1303}* & \textbf{0.1767}* &\textbf{\underline{0.8911}} & \textbf{0.1603}* & \textbf{0.0763}* \\ \hline

\multirow{6}{*}{100} & FedAvg & 0.8444* & 0.2145 & 0.1624 & 0.6889* & 0.3067 & {0.2948} & 0.9145* & 0.2799 & {0.1384} \\

& AFL & 0.8109 & 0.1422 & 0.1553 & 0.6293 & 0.2273 & \underline{0.2190} & 0.8352 & \underline{0.1908} & 0.0974  \\

& LocalFair & 0.8247 & 0.1592 & 0.1547 & 0.6336 & 0.2765 & 0.2547 & 0.8825 & 0.2034 & 0.1365 \\

& FEDFB & {0.8226} & 0.1463 & 0.1661 &  0.6360 & 0.2709 & 0.2667 & 0.8784 & 0.1972 & \underline{0.0896}  \\

& \MLocal & 0.8191 & \underline{0.1121} & \underline{0.1482} & {0.6399} & \underline{0.1985} & 0.2273 & 0.8872 & 0.2041 & 0.0914 \\

& \M & \textbf{\underline{0.8258}} & \textbf{0.1034*} & \textbf{{0.1314*}} & \underline{\textbf{0.6434}} & \textbf{0.1329}* & \textbf{0.1793}* & \textbf{\underline{0.8923}} & \textbf{0.1574}* & \textbf{0.0845}*   \\ \hline

\end{tabular}
}

\begin{tablenotes}
        \footnotesize
        \item[*] Note that the bold text indicates the result of our proposed \M~framework. The best results are marked with *. The second-best results are underlined.
        All outcomes pass the significance test, with a p-value below the significance threshold of $0.05$.
      \end{tablenotes}
    \end{threeparttable}
\end{table*}

\begin{table*}[!ht]
\centering
\renewcommand{\arraystretch}{1.1}
\caption{Experimental results with two sensitive groups.}
\label{experiment-result-two-attribute-table}
\begin{threeparttable}
\resizebox{\linewidth}{!}{
\begin{tabular}{clccccccccc}
\hline
& Dataset & \multicolumn{3}{c}{Adult} & \multicolumn{3}{c}{Compas} & \multicolumn{3}{c}{CelebA}  \\
\cmidrule(lr){3-5}\cmidrule(lr){6-8}\cmidrule(lr){9-11}

$\alpha$ & Method & Acc ($\uparrow$) & $\mathcal{M}_{DP}$ ($\downarrow$) & $\mathcal{M}_{EOP}$ ($\downarrow$) & Acc ($\uparrow$) & $\mathcal{M}_{DP}$ ($\downarrow$) & $\mathcal{M}_{EOP}$ ($\downarrow$)  & Acc ($\uparrow$) & $\mathcal{M}_{DP}$ ($\downarrow$) & $\mathcal{M}_{EOP}$ ($\downarrow$) \\ \hline
\multirow{4}{*}{0.1} & FedAvg & {0.8352*}   & 0.1624 & 0.1503 & 0.6741*   & 0.2569 & 0.2364 & 0.8989*   & 0.1076 & 0.0623 \\ 

& FADE & 0.8175 & 0.1056 & 0.0914 & \underline{0.6677} & 0.2316 & 0.2032 & 0.8822 & 0.0914 & 0.0488 \\

& FairFed & 0.8152 &  \underline{0.1033} &  \underline{0.0835} & 0.6455 &  \underline{0.1812} & \underline{0.1417} & 0.8731 & \underline{0.0856}  &  {0.0390*} \\

& \M & \textbf{ \underline{0.8240}} & \textbf{0.0860*} & \textbf{0.0322*} & \textbf{0.6498} & \textbf{0.1567*}   & \textbf{0.1344*}   & \textbf{\underline{0.8863}} & \textbf{ {0.0794*}} & \textbf{\underline{0.0417}}   \\ \hline

\multirow{4}{*}{0.5} & FedAvg & 0.8356*   & 0.1673 & 0.1515 & 0.6846*   & 0.2641 & 0.2483 & 0.9068*   & 0.1118 & 0.0608 \\ 

& FADE & 0.8199 & 0.1522 & 0.0521 & \underline{0.6500} & \underline{0.1892} & \underline{0.1913} & 0.8935 & 0.1003 & 0.0492 \\

& FairFed & 0.8183 &  \underline{0.1473} &  \underline{0.0464} & 0.6417 &  {0.1944} & 0.1992 & 0.8925 & \underline{0.0961}  &  \underline{0.0488} \\

& \M & \textbf{ \underline{0.8257}} & \textbf{0.0846*} & \textbf{0.0357*} & \textbf{0.6436} & \textbf{0.1329*}   & \textbf{0.1842*}   & \textbf{\underline{0.9002}} & \textbf{ {0.0807*}} & \textbf{0.0210*}   \\ \hline

\multirow{4}{*}{5} & FedAvg & 0.8372*   & 0.1651 & 0.1520 &  {0.6875*} & 0.2648 & 0.2502 &  {0.9070*} & 0.1120 & 0.0584  \\

& FADE & \underline{0.8302} & 0.1423 & 0.1366 & 0.6627 & 0.2334 & 0.2085 & 0.8834 & 0.1052 & 0.0496 \\

& FairFed & 0.8259 & \underline{0.1234}   & \underline{0.0973}   & 0.6622 & \underline{0.2055} & \underline{0.1978} & 0.8647 & \underline{0.0862} & \underline{0.0303}   \\

& \M & \textbf{0.8226} & \textbf{ {0.0877*}} & \textbf{ {0.0768*}} & \textbf{\underline{0.6661}} & \textbf{0.1695*}   & \textbf{0.1871*}   & \textbf{\underline{0.8919}}   & \textbf{0.0846*}   & \textbf{0.0200*}   \\ \hline

\multirow{4}{*}{20} & FedAvg & 0.8391*   & 0.1634 & 0.1531 &  {0.6879*} & 0.2652 & 0.2544 &  {0.9099*} & 0.1135 & 0.0566  \\

& FADE & \underline{0.8333} & 0.1457 & 0.1343 & \underline{0.6532} & 0.2201 & \underline{0.1967} & 0.8924 & 0.1055 & 0.0440 \\

& FairFed & 0.8289 & \underline{0.1277}   & \underline{0.1031}   & 0.6451 & \underline{0.2091} & 0.2001 & 0.8863 & \underline{0.0921} & \underline{0.0352}   \\

& \M & \textbf{0.8258} & \textbf{ {0.1054*}} & \textbf{ {0.0899*}} & \textbf{0.6491} & \textbf{0.1884*}   & \textbf{0.1890*}   & \textbf{\underline{0.8941}}   & \textbf{0.0855*}   & \textbf{0.0189*}   \\ \hline

\multirow{4}{*}{100} & FedAvg & 0.8406*   & 0.1651 & 0.1551 & 0.6880*   & 0.2663 &  {0.2603} & 0.9171*   & 0.1188 &  {0.0439} \\

& FADE & \underline{0.8376} & 0.1504 & 0.1303 & 0.6320 & \underline{0.2015} & \underline{0.1814} & \underline{0.9056} & 0.1066 & 0.0404 \\

& FairFed & 0.8336 & \underline{0.1364}  & \underline{0.1142}   & 0.6370 & 0.2174 & 0.1895 & 0.9009 &  \underline{0.1008} & \underline{0.0374} \\

& \M & \textbf{0.8355} & \textbf{0.1232*} & \textbf{ {0.0990*}} & \textbf{\underline{0.6403}} & \textbf{0.1590*}   & \textbf{0.1762*}   & \textbf{ {0.9016}} & \textbf{0.0865*}   & \textbf{0.0108*}     \\ \hline
\end{tabular}
}
\begin{tablenotes}
        \footnotesize
        \item[*] Note that the bold text indicates the result of our proposed \M~framework. The best results are marked with *. The second-best results are underlined.
        All outcomes pass the significance test, with a p-value below the significance threshold of $0.05$.
      \end{tablenotes}
\end{threeparttable}
\end{table*}


\begin{figure*}[ht]
    \centering    \includegraphics[width=0.95\linewidth]{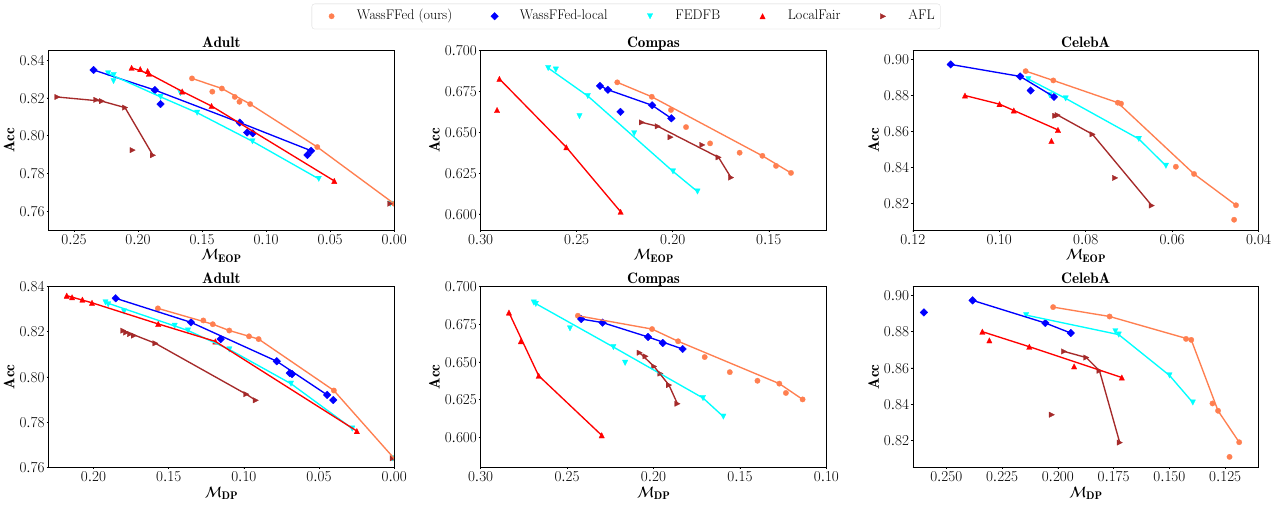}
    \caption{The Pareto frontier of $\mathcal{M}_{EOP}$ and $\mathcal{M}_{DP}$ on Compas, Adult, and CelebA datasets. The curve closer to the upper right corner indicates a better trade-off between accuracy and fairness.}
    \label{pareto-eop}
\end{figure*}

\begin{figure*}[ht]
    \centering
    \includegraphics[width=0.95\linewidth]{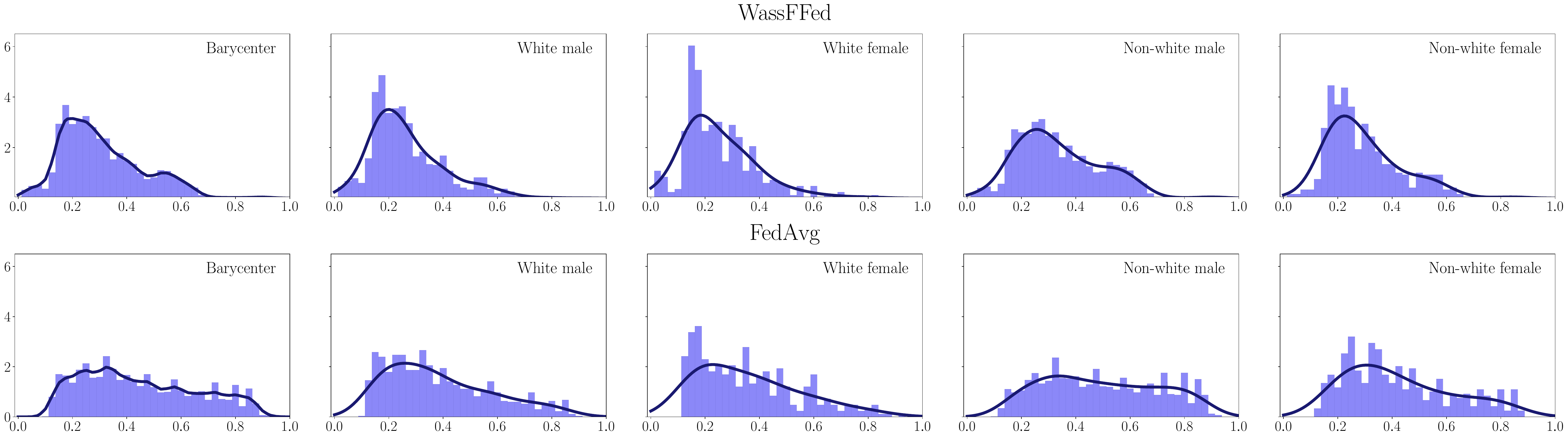}
    \caption{This figure demonstrates the model output distributions for various sensitive groups on the Compas dataset. The results provide evidence that \M~successfully achieves a model whose outputs are independent of sensitive attributes.}
    \label{distributions-figure}
\end{figure*}
\nosection{Baselines}
Research on fairness in FL is still in its early stages.
Some approaches can be generalized to handle multiple sensitive groups, while others are restricted to scenarios involving only two sensitive groups.
Our proposed method, \M, is designed to address fairness issues in both multi-sensitive and two-sensitive group settings.
To thoroughly assess the effectiveness of \M, we conduct experiments comparing it against methods tailored to both multi- and two-sensitive group scenarios.
We utilize FedAvg~\cite{mcmahan2017communication} as the optimal accuracy benchmark in fairness under FL, which introduces the average aggregation approach in the context of FL.
\textit{It should not be compared directly with fairness-oriented baselines unless those baselines perform worse in both accuracy and fairness compared with FedAvg.}

\textit{For the setting of multi-sensitive group}, we categorize all datasets into four sensitive groups, including non-white male, white male, non-white female, and white female.
The comparable methods are as follows:
\begin{itemize}
    \item \textbf{AFL}~\cite{mohri2019agnostic}, a SOTA method that defines an agnostic and more risk-averse objective to deal with any possible target distribution formed by a mixture of client distributions.
    \item \textbf{FEDFB}~\cite{zeng2021improving}, a SOTA method that presents a FairBatch-based approach~\cite{roh2020fairbatch} to compute the coefficients of FairBatch parameters on the server.
    \item \textbf{LocalFair} which trains the FairBatch model on each client and subsequently aggregates the parameters of local models following the FedAvg framework.
    \item \textbf{\MLocal}, which calculates the Wasserstein barycenter on each client and then aggregates the parameters of local models, also in accordance with the FedAvg methodology. This method is designed for ablation study.
\end{itemize}
Through the comparison of LocalFair and \MLocal~with \M, we aim to demonstrate that simply aggregating local fair models may not necessarily lead to the attainment of a globally fair model, highlighting the importance of addressing CH2, as introduced in Section~\ref{sec:introduction}.
The experimental results are sumarrized in Table~\ref{multi-sensitive-experiment-result-table}.

\textit{For the setting of two-sensitive groups}, following existing research~\cite{le2022survey}, we select gender (male and female) as the sensitive attribute in Adult, and race (white and non-white) in Compas and CelebA.
The comparable methods are as follows:
\begin{itemize}
    \item \textbf{FADE}~\cite{hong2021federated}, a SOTA method that introduces a federated adversarial debiasing method to attain the same global optimality as the one by the central algorithm.
    \item \textbf{FairFed}~\cite{ezzeldin2023fairfed} a SOTA method that is server-side and agnostic to the applied local debiasing thus allowing for ﬂexible use of different local debiasing methods across clients.
\end{itemize}
The experimental results are summarized in Table~\ref{experiment-result-two-attribute-table}.

\nosection{Evaluation Protocols}
\textit{Firstly}, we partition each dataset into an 70\% training set and reserve the remaining 30\% for testing.
\textit{Secondly}, we create a distribution of users in each sensitive group for every client, following the Dirichlet distribution $Dir(\alpha)$~\cite{wang2023mitigating}. A larger value of $\alpha$ indicates greater client homogeneity.
\textit{Thirdly}, we set the number of clients as 4 to better simulate the essential non-IID data distributions of four sensitive groups.
\textit{Fourthly}, we evaluate the FL model with Accuracy (Acc), $\mathcal{M}_{DP}$, and $\mathcal{M}_{EOP}$. Smaller values of $\mathcal{M}_{DP}$ and $\mathcal{M}_{EOP}$ denote a fairer model.
We run each model 5 times in each dataset and save the average performance.

\nosection{Parameter Settings}
\textit{For \M}, we configure the hyperparameter $\epsilon$ with a value of 1, as recommended in~\cite{cuturi2013sinkhorn}.
Besides, we set the value of $\lambda_{t}^{a}$ as $\frac{1}{|\mathcal{A}|}$ following~\cite{cuturi2014fast}.
We set the value of the trade-off hyperparameter $\beta$, the number of local rounds $k$, the number of bins $N_B$, and the differential privacy probability $\xi$ according to the experimental results of hyperparameters (see Section~\ref{hyperparameter-effect-section}).
\textit{For AFL, FEDFB, FADE, and FairFd}, we use the codes provided by authors and retain their default parameter settings.
To ensure a fair comparison, we employ the Adam optimizer~\cite{kingma2014adam} with a uniform learning rate of 0.005 across all models.
In addition, we establish the number of iteration rounds between the global model and clients as 50 to guarantee convergence.


\subsection{Overall Comparison (RQ1)}
We conduct extensive experiments on three public real-world datasets and report the experimental results with multi- and two-sensitive groups in Table~\ref{multi-sensitive-experiment-result-table} and Table~\ref{experiment-result-two-attribute-table}.
We also report the Pareto frontier in Figure~\ref{pareto-eop} to evaluate the ability to strike a balance between accuracy and fairness ($\mathcal{M}_{EOP}$ and $\mathcal{M}_{DP}$) for each fairness method.
Overall, in all three datasets with both multi- and two-sensitive group settings, when compared to existing SOTA methods, \M~consistently demonstrates the highest capability to strike a balance between accuracy and fairness.
The reason is that \M~has the ability to escape from the potential negative effect brought by the surrogate function (\textbf{CH1}) and achieve a better global fair model based on the guide from the global Wasserstein barycenter (\textbf{CH2}).

\subsection{Ablation Study (RQ2)}
To demonstrate the necessity of computing a global distribution for fairness, we design a model, \MLocal, which computes a Wasserstein barycenter on each client and aggregates local models following the FedAvg approach.
The results of this ablation study are presented in Table~\ref{multi-sensitive-experiment-result-table}.
It is evident that \M~consistently outperforms \MLocal~in terms of achieving a balance between accuracy and fairness.
The reason behind this is that \MLocal~solely focuses on training local fair models and aggregating them.
However, in many practical scenarios, the data distributions significantly deviate from the ideal non-IID situation across clients, resulting in incongruities between the local fair models and a global fair model.
In contrast, \M~computes a global distribution, specifically the global Wasserstein barycenter, and ensures that the output distributions of various sensitive groups from all clients are aligned with this global distribution.
This approach guarantees a consistent fair result shared between clients and the server, leading to the development of a superior fair global model.

\subsection{Output Distribution (RQ3)}
To assess whether \M~is capable of generating output distributions independent of sensitive attributes, we take the Compas dataset as an example, to visualize the output distributions with $\alpha = 0.5$ in Figure~\ref{distributions-figure}.
The results of FedAvg indicate that the output distributions for non-white-male users are notably distinct from those of other sensitive groups.
Non-white males are more likely to be predicted as potential re-offenders, which is evidently unfair.
In contrast, \M~successfully mitigates such unfairness.
Different sensitive groups share similar output distributions, independent of their sensitive attributes.
As presented in Table~\ref{multi-sensitive-experiment-result-table} and ~\ref{experiment-result-two-attribute-table}, this fair model can still maintain a high level of accuracy.

\subsection{Effect of Hyperparameters (RQ4) \label{hyperparameter-effect-section}}
We conducted experiments in \textbf{Appendix A} to demonstrate the effect of important hyperparameters for \M.

\subsection{Effect of the number of clients (RQ5) \label{effect-of-client-number-section}}
We conducted experiments in \textbf{Appendix B} to demonstrate the effect of the number of clients for \M.



\section{Conclusion}
In this paper, we introduce a novel framework called \textbf{Wass}erstein \textbf{F}air \textbf{Fed}erated Learning, denoted as \M, designed to ensure group fairness within the context of FL.
\M~achieves fairness by computing a global Wasserstein barycenter based on model output distributions across various sensitive groups from all clients.
It subsequently enforces the output distributions of users with distinct sensitive attributes within each client to align with this global barycenter.
This approach ensures that model outputs are independent of sensitive attributes, thereby yielding a fair model.
Besides, \M~circumvents the inherent estimation errors stemming from the utilization of surrogate functions and maintains consistency between the global fair model and client fairness.
We conduct extensive experiments on three publicly available real-world datasets.
Experimental results demonstrate that \M~outperforms SOTA methods.
It exhibits a remarkable ability to strike a harmonious balance between accuracy and fairness.

\ifCLASSOPTIONcaptionsoff
  \newpage
\fi



%
\bibliography{IEEEabrv, reference}

\begin{thebibliography}{10}
\providecommand{\url}[1]{#1}
\csname url@samestyle\endcsname
\providecommand{\newblock}{\relax}
\providecommand{\bibinfo}[2]{#2}
\providecommand{\BIBentrySTDinterwordspacing}{\spaceskip=0pt\relax}
\providecommand{\BIBentryALTinterwordstretchfactor}{4}
\providecommand{\BIBentryALTinterwordspacing}{\spaceskip=\fontdimen2\font plus
\BIBentryALTinterwordstretchfactor\fontdimen3\font minus \fontdimen4\font\relax}
\providecommand{\BIBforeignlanguage}[2]{{%
\expandafter\ifx\csname l@#1\endcsname\relax
\typeout{** WARNING: IEEEtran.bst: No hyphenation pattern has been}%
\typeout{** loaded for the language `#1'. Using the pattern for}%
\typeout{** the default language instead.}%
\else
\language=\csname l@#1\endcsname
\fi
#2}}
\providecommand{\BIBdecl}{\relax}
\BIBdecl

\bibitem{binns2018fairness}
R.~Binns, ``Fairness in machine learning: Lessons from political philosophy,'' in \emph{Conference on fairness, accountability and transparency}.\hskip 1em plus 0.5em minus 0.4em\relax PMLR, 2018, pp. 149--159.

\bibitem{han2023processing}
Z.~Han, C.~Chen, X.~Zheng, W.~Liu, J.~Wang, W.~Cheng, and Y.~Li, ``In-processing user constrained dominant sets for user-oriented fairness in recommender systems,'' \emph{arXiv preprint arXiv:2309.01335}, 2023.

\bibitem{hutchinson201950}
B.~Hutchinson and M.~Mitchell, ``50 years of test (un) fairness: Lessons for machine learning,'' in \emph{Proceedings of the conference on fairness, accountability, and transparency}, 2019, pp. 49--58.

\bibitem{mehrabi2021survey}
N.~Mehrabi, F.~Morstatter, N.~Saxena, K.~Lerman, and A.~Galstyan, ``A survey on bias and fairness in machine learning,'' \emph{ACM computing surveys (CSUR)}, vol.~54, no.~6, pp. 1--35, 2021.

\bibitem{dwork2012fairness}
C.~Dwork, M.~Hardt, T.~Pitassi, O.~Reingold, and R.~Zemel, ``Fairness through awareness,'' in \emph{Proceedings of the 3rd innovations in theoretical computer science conference}, 2012, pp. 214--226.

\bibitem{hardt2016equality}
M.~Hardt, E.~Price, and N.~Srebro, ``Equality of opportunity in supervised learning,'' \emph{Advances in neural information processing systems}, vol.~29, 2016.

\bibitem{zafar2017parity}
M.~B. Zafar, I.~Valera, M.~Rodriguez, K.~Gummadi, and A.~Weller, ``From parity to preference-based notions of fairness in classification,'' \emph{Advances in neural information processing systems}, vol.~30, 2017.

\bibitem{mcmahan2017communication}
B.~McMahan, E.~Moore, D.~Ramage, S.~Hampson, and B.~A. y~Arcas, ``Communication-efficient learning of deep networks from decentralized data,'' in \emph{Artificial intelligence and statistics}.\hskip 1em plus 0.5em minus 0.4em\relax PMLR, 2017, pp. 1273--1282.

\bibitem{smith2017federated}
V.~Smith, C.-K. Chiang, M.~Sanjabi, and A.~S. Talwalkar, ``Federated multi-task learning,'' \emph{Advances in neural information processing systems}, vol.~30, 2017.

\bibitem{yang2019federated}
Q.~Yang, Y.~Liu, T.~Chen, and Y.~Tong, ``Federated machine learning: Concept and applications,'' \emph{ACM Transactions on Intelligent Systems and Technology (TIST)}, vol.~10, no.~2, pp. 1--19, 2019.

\bibitem{du2021fairness}
W.~Du, D.~Xu, X.~Wu, and H.~Tong, ``Fairness-aware agnostic federated learning,'' in \emph{Proceedings of the 2021 SIAM International Conference on Data Mining (SDM)}.\hskip 1em plus 0.5em minus 0.4em\relax SIAM, 2021, pp. 181--189.

\bibitem{goh2016satisfying}
G.~Goh, A.~Cotter, M.~Gupta, and M.~P. Friedlander, ``Satisfying real-world goals with dataset constraints,'' \emph{Advances in neural information processing systems}, vol.~29, 2016.

\bibitem{menon2018cost}
A.~K. Menon and R.~C. Williamson, ``The cost of fairness in binary classification,'' in \emph{Conference on Fairness, accountability and transparency}.\hskip 1em plus 0.5em minus 0.4em\relax PMLR, 2018, pp. 107--118.

\bibitem{zafar2017fairness}
M.~B. Zafar, I.~Valera, M.~G. Rogriguez, and K.~P. Gummadi, ``Fairness constraints: Mechanisms for fair classification,'' in \emph{Artificial intelligence and statistics}.\hskip 1em plus 0.5em minus 0.4em\relax PMLR, 2017, pp. 962--970.

\bibitem{wu2019convexity}
Y.~Wu, L.~Zhang, and X.~Wu, ``On convexity and bounds of fairness-aware classification,'' in \emph{The World Wide Web Conference}, 2019, pp. 3356--3362.

\bibitem{lohaus2020too}
M.~Lohaus, M.~Perrot, and U.~Von~Luxburg, ``Too relaxed to be fair,'' in \emph{International Conference on Machine Learning}.\hskip 1em plus 0.5em minus 0.4em\relax PMLR, 2020, pp. 6360--6369.

\bibitem{abay2020mitigating}
A.~Abay, Y.~Zhou, N.~Baracaldo, S.~Rajamoni, E.~Chuba, and H.~Ludwig, ``Mitigating bias in federated learning,'' \emph{arXiv preprint arXiv:2012.02447}, 2020.

\bibitem{ezzeldin2023fairfed}
Y.~H. Ezzeldin, S.~Yan, C.~He, E.~Ferrara, and A.~S. Avestimehr, ``Fairfed: Enabling group fairness in federated learning,'' in \emph{Proceedings of the AAAI Conference on Artificial Intelligence}, vol.~37, no.~6, 2023, pp. 7494--7502.

\bibitem{wang2023mitigating}
G.~Wang, A.~Payani, M.~Lee, and R.~Kompella, ``Mitigating group bias in federated learning: Beyond local fairness,'' \emph{arXiv preprint arXiv:2305.09931}, 2023.

\bibitem{zeng2021improving}
Y.~Zeng, H.~Chen, and K.~Lee, ``Improving fairness via federated learning,'' \emph{arXiv preprint arXiv:2110.15545}, 2021.

\bibitem{vallender1974calculation}
S.~Vallender, ``Calculation of the wasserstein distance between probability distributions on the line,'' \emph{Theory of Probability \& Its Applications}, vol.~18, no.~4, pp. 784--786, 1974.

\bibitem{jiang2020wasserstein}
R.~Jiang, A.~Pacchiano, T.~Stepleton, H.~Jiang, and S.~Chiappa, ``Wasserstein fair classification,'' in \emph{Uncertainty in artificial intelligence}.\hskip 1em plus 0.5em minus 0.4em\relax PMLR, 2020, pp. 862--872.

\bibitem{dai2022comprehensive}
E.~Dai, T.~Zhao, H.~Zhu, J.~Xu, Z.~Guo, H.~Liu, J.~Tang, and S.~Wang, ``A comprehensive survey on trustworthy graph neural networks: Privacy, robustness, fairness, and explainability,'' \emph{arXiv preprint arXiv:2204.08570}, 2022.

\bibitem{corbett2017algorithmic}
S.~Corbett-Davies, E.~Pierson, A.~Feller, S.~Goel, and A.~Huq, ``Algorithmic decision making and the cost of fairness,'' in \emph{Proceedings of the 23rd acm sigkdd international conference on knowledge discovery and data mining}, 2017, pp. 797--806.

\bibitem{kusner2017counterfactual}
M.~J. Kusner, J.~Loftus, C.~Russell, and R.~Silva, ``Counterfactual fairness,'' \emph{Advances in neural information processing systems}, vol.~30, 2017.

\bibitem{berk2021fairness}
R.~Berk, H.~Heidari, S.~Jabbari, M.~Kearns, and A.~Roth, ``Fairness in criminal justice risk assessments: The state of the art,'' \emph{Sociological Methods \& Research}, vol.~50, no.~1, pp. 3--44, 2021.

\bibitem{grgic2016case}
N.~Grgic-Hlaca, M.~B. Zafar, K.~P. Gummadi, and A.~Weller, ``The case for process fairness in learning: Feature selection for fair decision making,'' in \emph{NIPS symposium on machine learning and the law}, vol.~1, no.~2.\hskip 1em plus 0.5em minus 0.4em\relax Barcelona, Spain, 2016, p.~11.

\bibitem{d2017conscientious}
B.~d'Alessandro, C.~O'Neil, and T.~LaGatta, ``Conscientious classification: A data scientist's guide to discrimination-aware classification,'' \emph{Big data}, vol.~5, no.~2, pp. 120--134, 2017.

\bibitem{kang2020inform}
J.~Kang, J.~He, R.~Maciejewski, and H.~Tong, ``Inform: Individual fairness on graph mining,'' in \emph{Proceedings of the 26th ACM SIGKDD International Conference on Knowledge Discovery \& Data Mining}, 2020, pp. 379--389.

\bibitem{dai2020learning}
E.~Dai and S.~Wang, ``Learning fair graph neural networks with limited and private sensitive attribute information,'' \emph{arXiv preprint arXiv:2009.01454}, 2020.

\bibitem{bose2019compositional}
A.~Bose and W.~Hamilton, ``Compositional fairness constraints for graph embeddings,'' in \emph{International Conference on Machine Learning}.\hskip 1em plus 0.5em minus 0.4em\relax PMLR, 2019, pp. 715--724.

\bibitem{mammen2021federated}
P.~M. Mammen, ``Federated learning: Opportunities and challenges,'' \emph{arXiv preprint arXiv:2101.05428}, 2021.

\bibitem{li2021ditto}
T.~Li, S.~Hu, A.~Beirami, and V.~Smith, ``Ditto: Fair and robust federated learning through personalization,'' in \emph{International Conference on Machine Learning}.\hskip 1em plus 0.5em minus 0.4em\relax PMLR, 2021, pp. 6357--6368.

\bibitem{lyu2020collaborative}
L.~Lyu, X.~Xu, Q.~Wang, and H.~Yu, ``Collaborative fairness in federated learning,'' \emph{Federated Learning: Privacy and Incentive}, pp. 189--204, 2020.

\bibitem{chang2023bias}
H.~Chang and R.~Shokri, ``Bias propagation in federated learning,'' \emph{arXiv preprint arXiv:2309.02160}, 2023.

\bibitem{cui2021addressing}
S.~Cui, W.~Pan, J.~Liang, C.~Zhang, and F.~Wang, ``Addressing algorithmic disparity and performance inconsistency in federated learning,'' \emph{Advances in Neural Information Processing Systems}, vol.~34, pp. 26\,091--26\,102, 2021.

\bibitem{papadaki2022minimax}
A.~Papadaki, N.~Martinez, M.~Bertran, G.~Sapiro, and M.~Rodrigues, ``Minimax demographic group fairness in federated learning,'' in \emph{Proceedings of the 2022 ACM Conference on Fairness, Accountability, and Transparency}, 2022, pp. 142--159.

\bibitem{mohri2019agnostic}
M.~Mohri, G.~Sivek, and A.~T. Suresh, ``Agnostic federated learning,'' in \emph{International Conference on Machine Learning}.\hskip 1em plus 0.5em minus 0.4em\relax PMLR, 2019, pp. 4615--4625.

\bibitem{roh2020fairbatch}
Y.~Roh, K.~Lee, S.~E. Whang, and C.~Suh, ``Fairbatch: Batch selection for model fairness,'' \emph{arXiv preprint arXiv:2012.01696}, 2020.

\bibitem{dunda2023handling}
G.~W.~M. Dunda and S.~Song, ``Handling group fairness in federated learning using augmented lagrangian approach,'' \emph{arXiv preprint arXiv:2307.04417}, 2023.

\bibitem{zhang2020fairfl}
D.~Y. Zhang, Z.~Kou, and D.~Wang, ``Fairfl: A fair federated learning approach to reducing demographic bias in privacy-sensitive classification models,'' in \emph{2020 IEEE International Conference on Big Data (Big Data)}.\hskip 1em plus 0.5em minus 0.4em\relax IEEE, 2020, pp. 1051--1060.

\bibitem{kairouz2021advances}
P.~Kairouz, H.~B. McMahan, B.~Avent, A.~Bellet, M.~Bennis, A.~N. Bhagoji, K.~Bonawitz, Z.~Charles, G.~Cormode, R.~Cummings \emph{et~al.}, ``Advances and open problems in federated learning,'' \emph{Foundations and Trends{\textregistered} in Machine Learning}, vol.~14, no. 1--2, pp. 1--210, 2021.

\bibitem{cuturi2014fast}
M.~Cuturi and A.~Doucet, ``Fast computation of wasserstein barycenters,'' in \emph{International conference on machine learning}.\hskip 1em plus 0.5em minus 0.4em\relax PMLR, 2014, pp. 685--693.

\bibitem{villani2009optimal}
C.~Villani \emph{et~al.}, \emph{Optimal transport: old and new}.\hskip 1em plus 0.5em minus 0.4em\relax Springer, 2009, vol. 338.

\bibitem{bogachev2012monge}
V.~I. Bogachev and A.~V. Kolesnikov, ``The monge-kantorovich problem: achievements, connections, and perspectives,'' \emph{Russian Mathematical Surveys}, vol.~67, no.~5, p. 785, 2012.

\bibitem{ruschendorf1985wasserstein}
L.~R{\"u}schendorf, ``The wasserstein distance and approximation theorems,'' \emph{Probability Theory and Related Fields}, vol.~70, no.~1, pp. 117--129, 1985.

\bibitem{cuturi2013sinkhorn}
M.~Cuturi, ``Sinkhorn distances: Lightspeed computation of optimal transport,'' \emph{Advances in neural information processing systems}, vol.~26, 2013.

\bibitem{dwork2014differential_privacy}
C.~Dwork, A.~Roth \emph{et~al.}, ``The algorithmic foundations of differential privacy,'' \emph{Foundations and Trends{\textregistered} in Theoretical Computer Science}, vol.~9, no. 3--4, pp. 211--407, 2014.

\bibitem{deshpande2018generative}
I.~Deshpande, Z.~Zhang, and A.~G. Schwing, ``Generative modeling using the sliced wasserstein distance,'' in \emph{Proceedings of the IEEE conference on computer vision and pattern recognition}, 2018, pp. 3483--3491.

\bibitem{asuncion2007uci}
A.~Asuncion and D.~Newman, ``Uci machine learning repository,'' 2007.

\bibitem{dressel2018accuracy}
J.~Dressel and H.~Farid, ``The accuracy, fairness, and limits of predicting recidivism,'' \emph{Science advances}, vol.~4, no.~1, p. eaao5580, 2018.

\bibitem{zhang2020celeba}
Y.~Zhang, Z.~Yin, Y.~Li, G.~Yin, J.~Yan, J.~Shao, and Z.~Liu, ``Celeba-spoof: Large-scale face anti-spoofing dataset with rich annotations,'' in \emph{Computer Vision--ECCV 2020: 16th European Conference, Glasgow, UK, August 23--28, 2020, Proceedings, Part XII 16}.\hskip 1em plus 0.5em minus 0.4em\relax Springer, 2020, pp. 70--85.

\bibitem{jiang2020identifying}
H.~Jiang and O.~Nachum, ``Identifying and correcting label bias in machine learning,'' in \emph{International Conference on Artificial Intelligence and Statistics}.\hskip 1em plus 0.5em minus 0.4em\relax PMLR, 2020, pp. 702--712.

\bibitem{lavalley2008logistic}
M.~P. LaValley, ``Logistic regression,'' \emph{Circulation}, vol. 117, no.~18, pp. 2395--2399, 2008.

\bibitem{he2016deep}
K.~He, X.~Zhang, S.~Ren, and J.~Sun, ``Deep residual learning for image recognition,'' in \emph{Proceedings of the IEEE conference on computer vision and pattern recognition}, 2016, pp. 770--778.

\bibitem{le2022survey}
T.~Le~Quy, A.~Roy, V.~Iosifidis, W.~Zhang, and E.~Ntoutsi, ``A survey on datasets for fairness-aware machine learning,'' \emph{Wiley Interdisciplinary Reviews: Data Mining and Knowledge Discovery}, vol.~12, no.~3, p. e1452, 2022.

\bibitem{hong2021federated}
J.~Hong, Z.~Zhu, S.~Yu, Z.~Wang, H.~H. Dodge, and J.~Zhou, ``Federated adversarial debiasing for fair and transferable representations,'' in \emph{Proceedings of the 27th ACM SIGKDD Conference on Knowledge Discovery \& Data Mining}, 2021, pp. 617--627.

\bibitem{kingma2014adam}
D.~P. Kingma and J.~Ba, ``Adam: A method for stochastic optimization,'' \emph{arXiv preprint arXiv:1412.6980}, 2014.

\end{thebibliography}
\bibliographystyle{IEEEtran}




%
\begin{IEEEbiography}[{\includegraphics[width=1in,height=1.25in,clip,keepaspectratio]{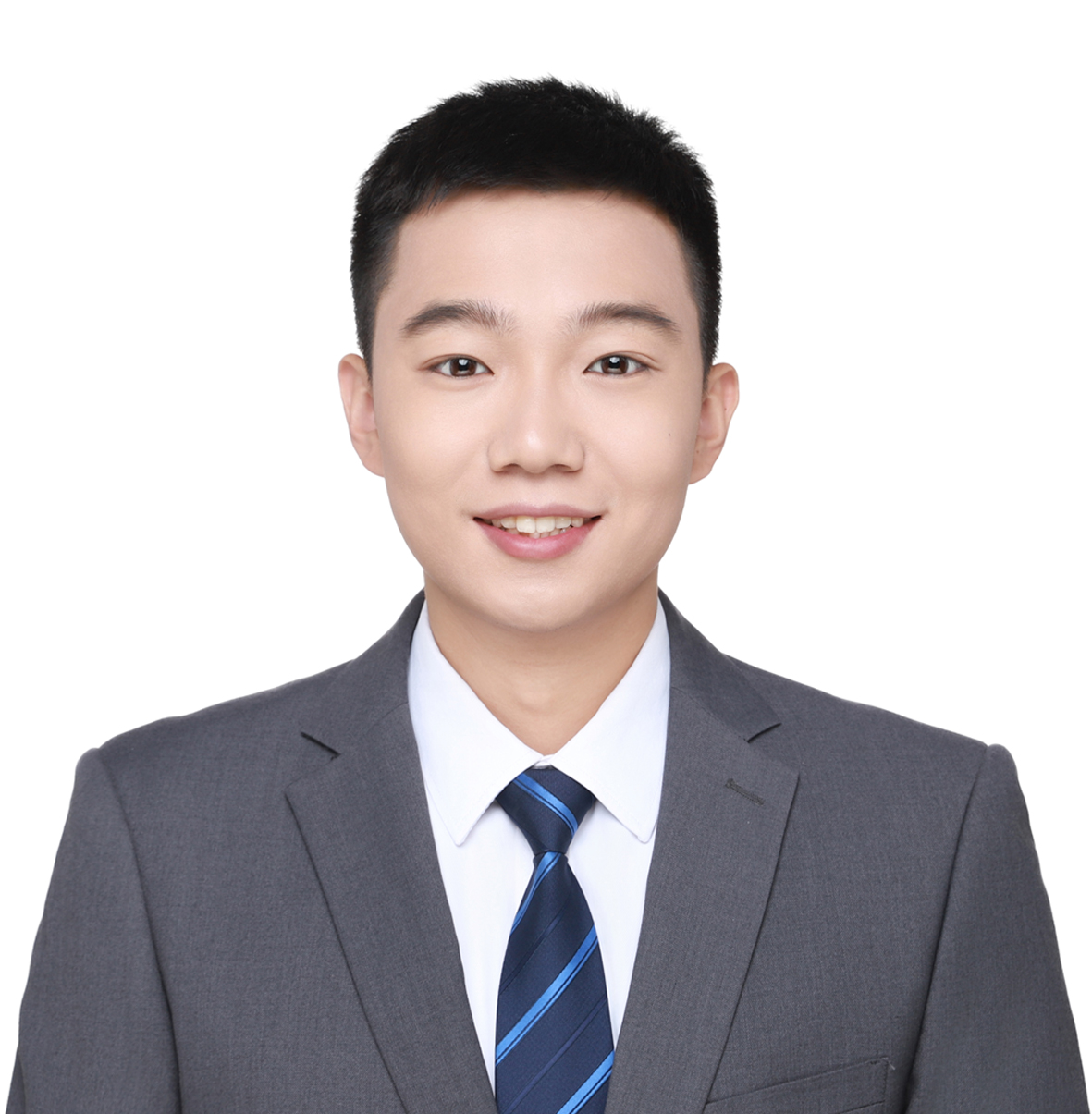}}]{Zhongxuan Han}
is currently pursuing a Ph.D. at the College of Computer Science and Technology, Zhejiang University. He graduated in 2020 with a Bachelor's degree from Chu Kochen Honors College, Zhejiang University. His research areas include recommender systems, machine learning fairness, and graph neural networks. He has published 8 papers in peer reviewed conferences such as ICML, SIGIR, WWW, AAAI, and ACM MM.
\end{IEEEbiography}

\begin{IEEEbiography}[{\includegraphics[width=1in,height=1.25in,clip,keepaspectratio]{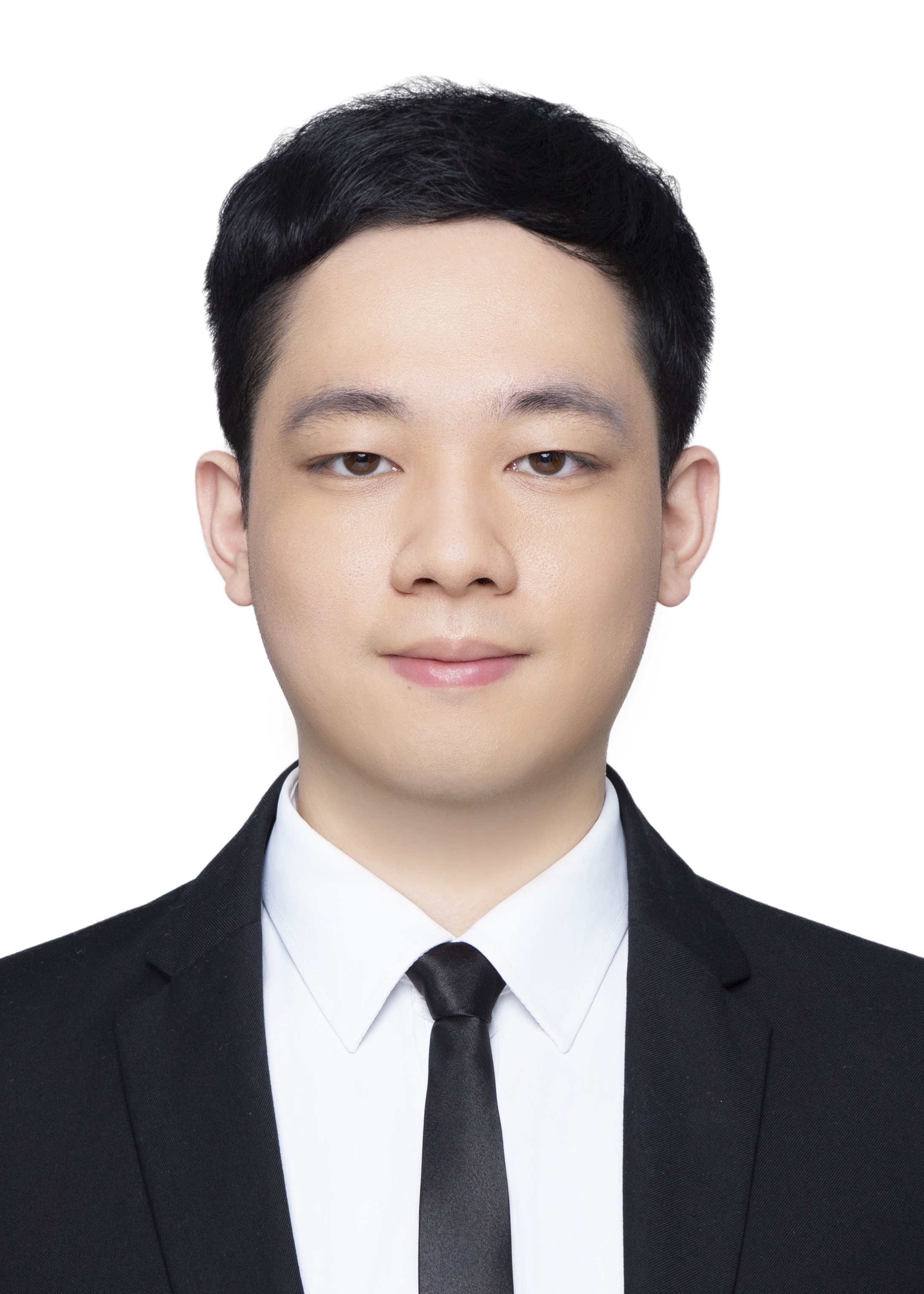}}]{Li Zhang}
obtained his Bachelor of Science in Statistics from Chongqing University, China, in 2022. He is currently pursuing a master degree in Electronic and Information Engineering at Zhejiang University. His research interests encompass machine learning and trustworthy artificial intelligence.
\end{IEEEbiography}

\begin{IEEEbiography}[{\includegraphics[width=1in,height=1.25in,clip,keepaspectratio]{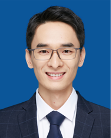}}]{Chaochao Chen}
obtained his PhD degree in computer science from Zhejiang University, China, in 2016, and he was a visiting scholar at the University of Illinois at Urbana-Champaign, during 2014-2015. He is currently a Distinguished Research Fellow at Zhejiang University. Before that, he was a Staff Algorithm Engineer at Ant Group. His research mainly focuses on recommender systems, privacy-preserving machine learning, and graph machine learning. He has published more than 80 papers in peer-reviewed journals and conferences.
\end{IEEEbiography}

\begin{IEEEbiography}[{\includegraphics[width=1in,height=1.25in,clip,keepaspectratio]{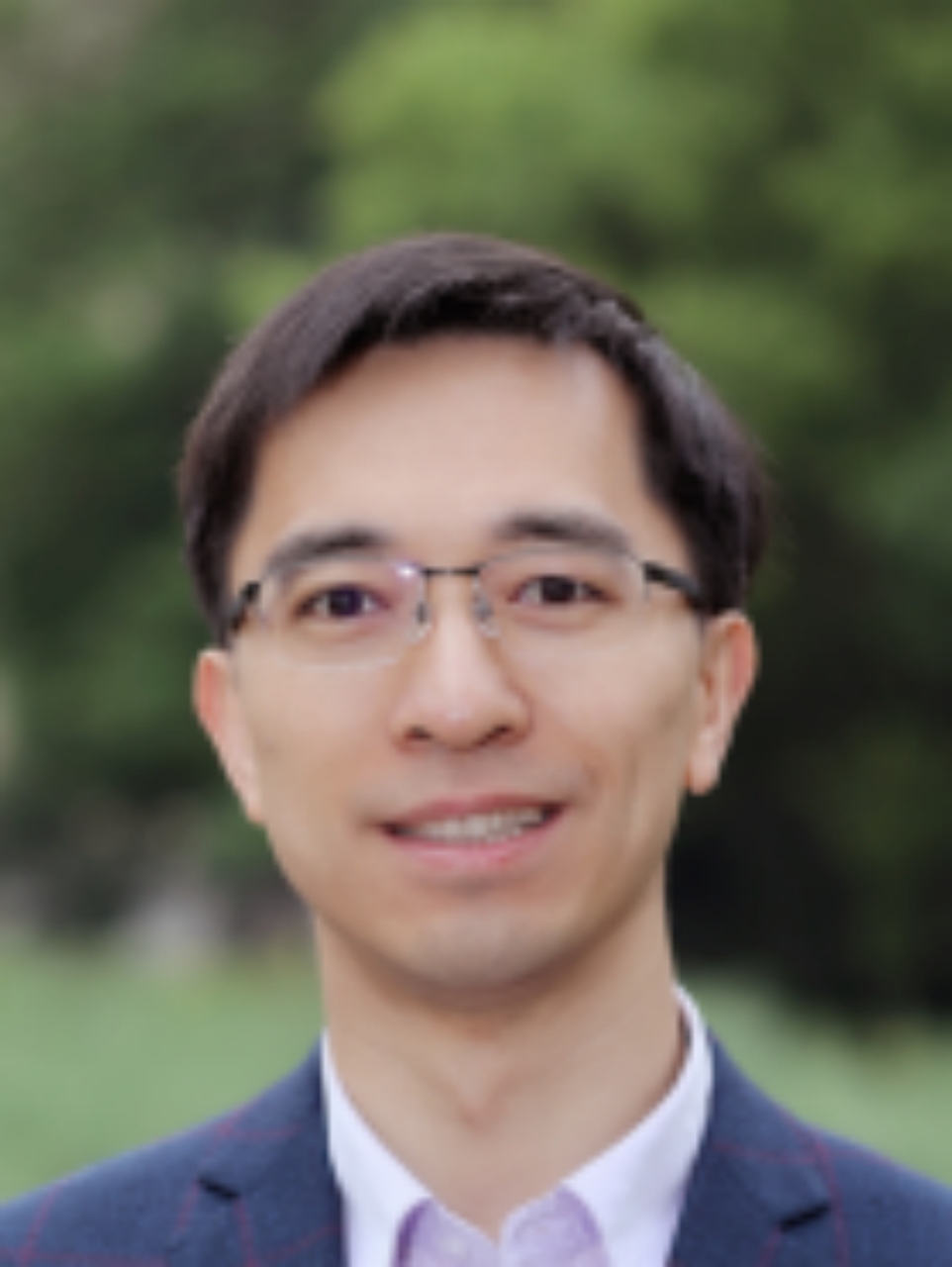}}]{Xiaolin Zheng}
PhD, Professor, PhD supervisor, and the deputy director of the Institute of Artificial Intelligence, Zhejiang University. Senior member of IEEE, and Distinguished Member of China Computer Federation, and a Committee Member in Service Computing of China Computer Federation. His main research interests include  Recommender Systems, Privacy-Preserving Computing, and Intelligent Finance. He has published more than 100 referenced papers in TKDE, NeurIPS, IJCAI, AAAI, WWW, KDD, MM and so on.
\end{IEEEbiography}

\begin{IEEEbiography}[{\includegraphics[width=1in,height=1.25in,clip,keepaspectratio]{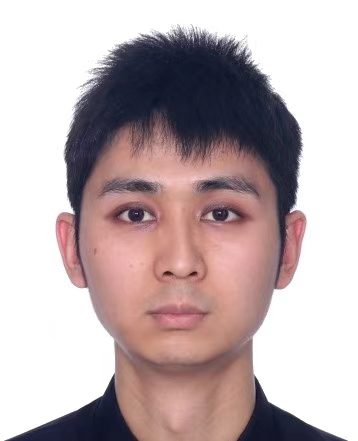}}]{Fei Zheng}
is currently pursuing a PhD degree at the College of Computer Science and Technology, Zhejiang University, Hangzhou, P.R. China. He received his bachelor's degree in Computer Science from the University of Science and Technology of China in 2019. His research interests include privacy-preserving machine learning and federated learning.
\end{IEEEbiography}

\begin{IEEEbiography}[{\includegraphics[width=1in,height=1.25in,clip,keepaspectratio]{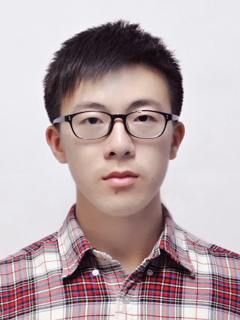}}]{Yuyuan Li}
obtained his PhD degree in computer science from Zhejiang University, China, in 2023. He is currently an Associate Professor at Hangzhou Dianzi University. His research interests mainly focus on trust-worthy machine learning. He has published more than 10 papers in peer reviewed journals and conferences, including NeurIPS, ICML, SIGIR, and CVPR.
\end{IEEEbiography}


\begin{IEEEbiography}[{\includegraphics[width=1in,height=1.25in,clip,keepaspectratio]{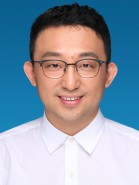}}]{Jianwei Yin}
received the PhD degree in computer science from Zhejiang University, in 2001. Senior member of IEEE. He is currently a professor with the College of Computer Science, Zhejiang University. He is a visiting scholar of Georgia Institute of Technology, in 2008. He has published more than 100 papers in top international journals and conferences. His research interests include service computing, software engineering and distributed computing.
\end{IEEEbiography}

\clearpage

\appendix

\subsection{Effect of Hyperparameters}
\begin{figure*}[t]
    \centering
    \subfigure[Adult]{
        \includegraphics[width=\textwidth]{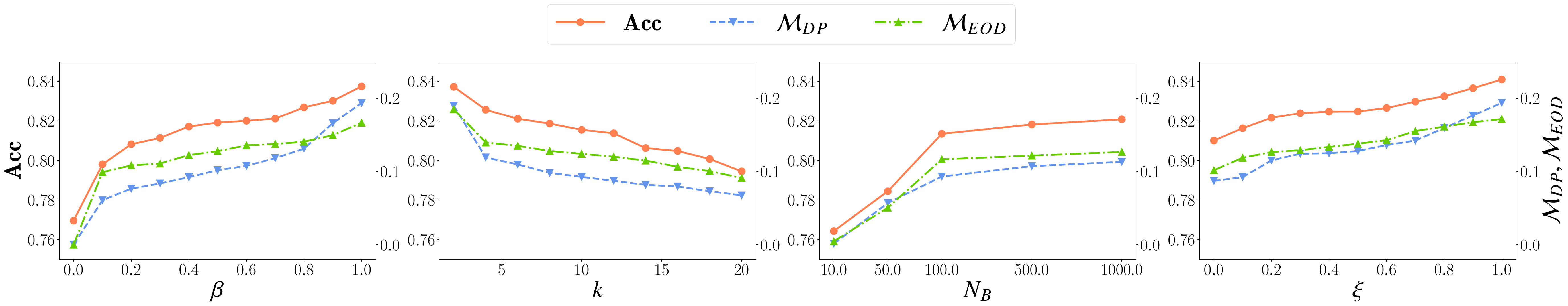}
    }
    \subfigure[Compas]{
        \includegraphics[width=\textwidth]{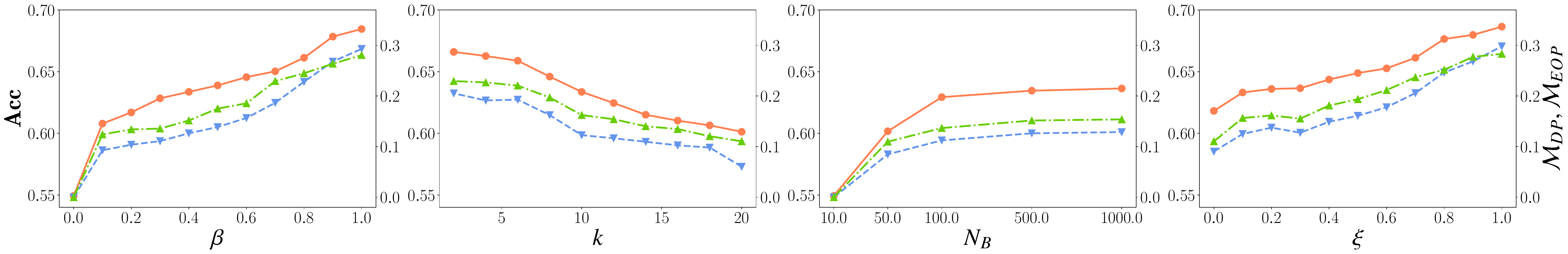}
    }
    \subfigure[CelebA]{
        \includegraphics[width=\textwidth]{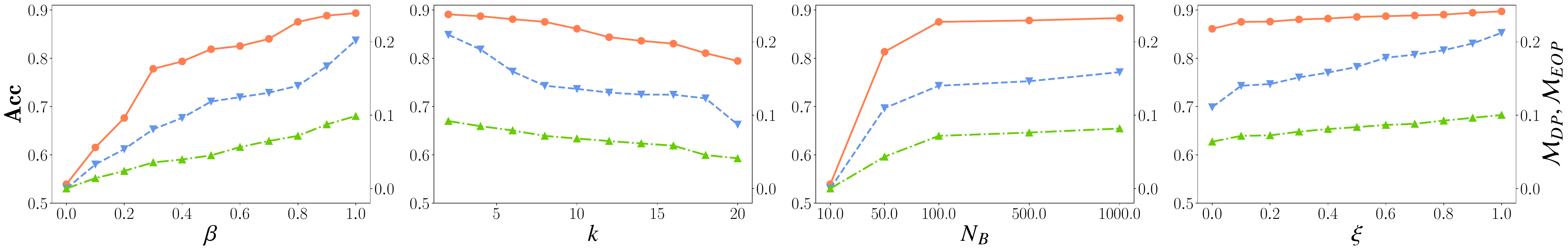}
    }
    \caption{Effect of hyperparameters.}
    \label{fig:hyperparameter}
\end{figure*}
We conducted experiments to investigate the effects of the trade-off hyperparameter $\beta$, the number of local rounds $k$, the number of bins $N_B$, and the differential privacy probability $\xi$.
The results are depicted in Figure~\ref{fig:hyperparameter}.

\nosection{Effect of $\beta$}
The value of $\beta$ determines the trade-off between fairness loss and model utility loss.
In Equation (14), as the value of $\beta$ increases, the model utility loss becomes more prominent while the fairness loss decreases.
This trend is consistent with the observations in experimental results, where an increase in the value of $\beta$ results in a more accurate yet less fair model.
When $\beta$ is set at 0.4, the model attains a balance between accuracy and fairness in all datasets.

\nosection{Effect of $k$}
The value of $k$ determines the number of local training rounds.
In our setup, when clients receive the optimal transport matrices from the server, they conduct local training for $k$ rounds.
As the value of $k$ increases, the model tends to become fairer, as the output distributions of various sensitive groups become more similar to the barycenter.
However, overly similar distributions can harm the model's accuracy, as it may overlook the unique characteristics of each group.
We set the value of $k$ to 15 since, in this setting, the model attains a satisfactory level of fairness while still maintaining a high degree of accuracy.

\nosection{Effect of $N_B$}
The parameter $N_B$ represents the number of bins utilized in processing the outputs of client models.
A value of $N_B$ that is too low results in a coarse approximation of the Wasserstein barycenter, potentially derailing the model from its standard optimization trajectory in pursuit of exaggerated fairness.
Conversely, a higher $N_B$ value enables a more precise estimation of the Wasserstein barycenter, facilitating a more effective equilibrium between fairness and accuracy.
Considering both model performance and privacy concerns, we select an $N_B$ value of $100$.

\nosection{Effect of $\xi$}
The parameter $\xi$ quantifies the level of differential privacy applied.
An increase in $\xi$ leads to more distorted client model outputs, which complicates the achievement of fairness objectives.
To strike a prudent balance between safeguarding client privacy and ensuring a satisfactory compromise between accuracy and fairness, we determine a $\xi$ value of $0.15$.

\subsection{Effect of the number of clients}

\begin{table}[!t]
\centering 
\renewcommand{\arraystretch}{1.1}
\caption{The effect of the number of clients on Adult}
\label{effect-client-table-adult}
\begin{threeparttable}
\resizebox{\linewidth}{!}{
\begin{tabular}{lcccccc}
\hline
& \multicolumn{3}{c}{\M} & \multicolumn{3}{c}{FedAvg} \\
\cmidrule(lr){2-4}\cmidrule(lr){5-7}
Clients & Acc ($\uparrow$) & $\mathcal{M}_{DP}$ ($\downarrow$) & $\mathcal{M}_{EOP}$ ($\downarrow$) & Acc ($\uparrow$) & $\mathcal{M}_{DP}$ ($\downarrow$) & $\mathcal{M}_{EOP}$ ($\downarrow$) \\ \hline 
2 & 0.8250 & 0.1374 & 0.1753 & 0.8312 & 0.2117 & 0.2040 \\ 
5 & 0.8254 & 0.1373 & 0.1949 & 0.8367  & 0.2093 & 0.1722 \\ 
10 & 0.8165 & 0.1124 & 0.1760 & 0.8311  & 0.2087 & 0.1739 \\
20 & 0.8041 & 0.0713 & 0.1226 & 0.8294  & 0.2101 & 0.1729\\
50 & 0.8037 & 0.0688 & 0.1162 & 0.8183   & 0.2146 & 0.1740 \\
100 & 0.8064 & 0.0764 &  0.1354 & 0.8155  & 0.2084 & 0.1687 \\ 

\hline
\end{tabular}
}

\end{threeparttable}
\end{table}

\begin{table}[!t]
\centering 
\renewcommand{\arraystretch}{1.1}
\caption{The effect of the number of clients on Compas}
\label{effect-client-table-compas}
\begin{threeparttable}
\resizebox{\linewidth}{!}{
\begin{tabular}{lcccccc}
\hline
& \multicolumn{3}{c}{\M} & \multicolumn{3}{c}{FedAvg} \\
\cmidrule(lr){2-4}\cmidrule(lr){5-7}
Clients & Acc ($\uparrow$) & $\mathcal{M}_{DP}$ ($\downarrow$) & $\mathcal{M}_{EOP}$ ($\downarrow$) & Acc ($\uparrow$) & $\mathcal{M}_{DP}$ ($\downarrow$) & $\mathcal{M}_{EOP}$ ($\downarrow$) \\ \hline  
2 & 0.6612 & 0.2382 &  0.2184 & 0.6884  & 0.3055 & 0.2913 \\ 
5 & 0.6427 & 0.2154 & 0.1947 & 0.6878   & 0.3013 & 0.2829 \\
10 & 0.6446 & 0.2178 & 0.1911 & 0.6775  & 0.2891 & 0.2692 \\
20 & 0.6332 & 0.1625 & 0.1684 & 0.6731  & 0.2880 & 0.2608 \\
50 & 0.6303 & 0.1589 & 0.1607 & 0.6698  & 0.2781 & 0.2593 \\ 
100 & 0.6241 & 0.1486 & 0.1415 & 0.6670 & 0.2610 & 0.2549 \\ 

\hline
\end{tabular}
}
\end{threeparttable}
\end{table}

\begin{table}[!t]
\centering 
\renewcommand{\arraystretch}{1.1}
\caption{The effect of the number of clients on Celeba}
\label{effect-client-table-celeba}
\begin{threeparttable}
\resizebox{\linewidth}{!}{
\begin{tabular}{lcccccc}
\hline
& \multicolumn{3}{c}{\M} & \multicolumn{3}{c}{FedAvg} \\
\cmidrule(lr){2-4}\cmidrule(lr){5-7}
Clients & Acc ($\uparrow$) & $\mathcal{M}_{DP}$ ($\downarrow$) & $\mathcal{M}_{EOP}$ ($\downarrow$) & Acc ($\uparrow$) & $\mathcal{M}_{DP}$ ($\downarrow$) & $\mathcal{M}_{EOP}$ ($\downarrow$) \\ \hline  
2 & 0.8988 & 0.1635 &  0.0815 & 0.9115  & 0.2749 & 0.1138 \\ 
5 & 0.9010 & 0.2354 & 0.0613 & 0.9035   & 0.2630 & 0.1039 \\
10 & 0.9016 & 0.1949 & 0.1091 & 0.8960  & 0.2301 & 0.1032 \\
20 & 0.8925 & 0.1730 & 0.0723 & 0.8957  & 0.2343 & 0.1017 \\
50 & 0.8837 & 0.1887 & 0.0815 & 0.8896  & 0.2201 & 0.1033 \\ 
100 & 0.8713 & 0.1487 & 0.0786 & 0.8825  & 0.2093 & 0.0941  \\ 

\hline
\end{tabular}
}

    \end{threeparttable}
\end{table}

To investigate the impact of the number of clients, we conduct experiments with $\alpha = 0.5$ with client numbers varying from $2, 5, 10, 20, 50$ to $100$.
The results of these experiments are depicted in Tables~\ref{effect-client-table-adult}, \ref{effect-client-table-compas}, and \ref{effect-client-table-celeba}. 
The findings indicate that as the number of clients increases, accuracy tends to decline, while fairness improves.

This reduction in accuracy is primarily due to the increased data heterogeneity accompanying a larger client base, complicating the aggregation process for a global model.
The \M~framework is designed to enhance fairness with minimal compromise on accuracy.
Across various client numbers, \M~manages to sustain high accuracy levels, experiencing only a slight reduction when compared to FedAvg.

Notably, although fairness in both \M~and FedAvg diminishes as the number of clients grows, \M~consistently outperforms FedAvg in terms of fairness.
This is attributed to the fact that a decrease in client sample size amplifies the impact of the global Wasserstein barycenter computed by \M, facilitating the achievement of overall fairness more effectively.





\end{document}